\documentclass[journal]{IEEEtran}
\usepackage{cite}
\usepackage{url}

\usepackage{microtype}
\usepackage{graphicx}
\usepackage{subfigure}
\usepackage{booktabs}
\usepackage{hyperref}
\usepackage{times}
\usepackage{epsfig}
\usepackage{graphicx}
\usepackage{amsmath}
\usepackage{amssymb}
\usepackage{placeins}
\usepackage{subfiles}
\usepackage{amsthm}
\usepackage{algorithm}
\usepackage{algorithmic}
\usepackage{pbox}
\usepackage{multirow}
\usepackage{float}

\DeclareMathOperator{\id}{id}
\DeclareMathOperator{\relu}{ReLU}
\DeclareMathOperator{\E}{\mathbb{E}}
\DeclareMathOperator{\leakyrelu}{LeakyReLU}

\begin{document}
\title{Ground Truth Free Denoising by Optimal Transport}

\author{S\"oren Dittmer,
        Carola-Bibiane Sch\"onlieb,
        and~Peter Maass
\thanks{S. Dittmer and P. Maass are with the University of Bremen.}
\thanks{C.-B. Sch\"onlieb is with the University of Cambridge.}
\thanks{Manuscript received July ??, 2020; revised ???? ??, 2020.}
\thanks{Accompanying code: \url{github.com/sdittmer/gtfd}}}

% The paper headers
\markboth{}%
{Dittmer \MakeLowercase{\textit{et al.}}: Ground Truth Free Denoising by Optimal Transport}

%\IEEEpubid{0000--0000/00\$00.00~\copyright~2015 IEEE}
\maketitle

\begin{abstract}
We present a learned unsupervised denoising method for \textbf{arbitrary types of data}, which we explore on images and one-dimensional signals. \textbf{The training is solely based on samples of noisy data and examples of noise}, which -- critically -- do not need to come in pairs. We only need the assumption that the noise is independent and additive (although we describe how this can be extended). The method rests on a Wasserstein Generative Adversarial Network setting, which utilizes two critics and one generator.
\end{abstract}

\begin{IEEEkeywords}
Deep learning, Optimal transport, Denoising
\end{IEEEkeywords}

\IEEEpeerreviewmaketitle

\section{Introduction}
Like many other fields, the field of inverse problems and denoising is undergoing major changes with the rise of deep learning. However, while deep learning methods provide an unprecedented improvement in reconstructions across the board~\cite{adler2018learned, jin2017deep, hammernik2018learning, zhang2017beyond, ye2018deep}, they heavily rely on the -- in practice often sparse  --  availability of training data. In particular, one often trains reconstructing networks based on sample pairs $(x, y^\delta)\sim p_{x, y^\delta}$, where the $x$ are given ground truth reconstructions of the noisy measurement $y^\delta$. This presupposes the availability of such pairs, which usually relies on either the capability to simulate such data points accurately or having already solved the inverse problem by other means (e.g., with high dosages in computer tomography~\cite{adler2018learned}).

Recently, there has been a flurry of papers considering the particular task of denoising without any ground truth data; see, e.g.,~\cite{soltanayev2018training, cha2018neural, metzler2018unsupervised, lehtinen2018noise2noise}. And although all these papers do not rely on ground truth data, they all seem to make one or several of the following assumptions, of which we make none: spacial independence of the noise, zero-centeredness of the noise, that the data possesses spacial structure, that they deal with Gaussian noise or that they have multiple noisy samples of the same underlying ground truth.

\begin{figure}[ht]
    \centering
    \includegraphics[trim={5cm 18.5cm 5cm 4cm},clip,width=.45\textwidth]{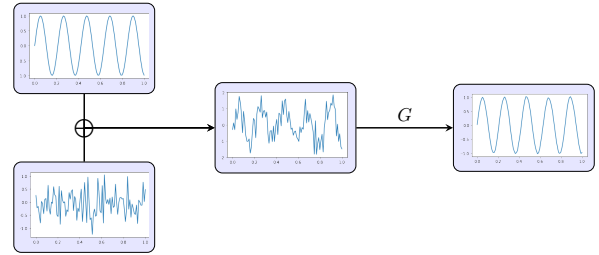}
    \caption{The measurement-reconstruction pipeline, displaying the measurement process followed by a denoising.}
    \label{fig:example_application_graph}
\end{figure}

\textbf{This paper aims at training a denoiser $G:Y\to Y$ with as easily available data as possible.} Specifically, we consider random noisy measurements $y^\delta$ from a probability density $p_{y^\delta}$ over a space $Y=\mathbb{R}^m$, which are generated by a probabilistic measurement process via adding arbitrary independent noise $\eta\sim p_\eta$ to clean data $y\sim p_y$, see Figure~\ref{fig:example_application_graph}.

Note, this means we only need a noisy data set $\{y^\delta_i\}_i$, and a noise data set $\{\eta_i\}_i$. To require the former seems highly unrestrictive, to request the later does often not exceed this, since many measurement operators are linear and therefore measuring noise equates to measuring the zero-element, e.g., in the case of computer tomography, measuring with an empty machine.
\FloatBarrier
We tackle the problem by making the following simple observations about an assumed ideal denoiser $G^*:Y\to Y$:
\begin{enumerate}
    \setcounter{enumi}{-1} 
    \item For a noisy measurement $y^\delta$, $G^*(y^\delta)$ is a clean data point.
    \item For additional arbitrary noise $\eta$, $G^*(y^\delta)+\eta$ looks like a noisy measurement.
    \item $y^\delta - G^*(y^\delta)$ looks like noise.
\end{enumerate}
We utilize these observations via an optimal transport formulation realized by a WGAN~\cite{arjovsky2017wasserstein} (Wasserstein generative adversarial network) setting, which consists of two critics to capitalize on point 1 and 2 of the enumeration above.

The paper is structured as follows. Subsequent to a related work subsection, we start with a reminder on ``classical'' WGANs, we then move on to the rigorous mathematical formulation of the enumeration above and how these observations can be cast into a training setup for a denoiser $G$. Subsequently to this, we present the numerical example of denoising highly noisy measurements of sines. We follow up with numerical experiments on STL-10 images~\cite{coates2011analysis} for Gaussian and non-Gaussian denoising, as well as on denoising of noisy and blurry measurements and subsequent deblurring by total variation~\cite{rudin1992nonlinear}. Before we conclude the paper, we outline how the method could be generalized to multiplicative noise.

\subsection{Related work}
Sparked by the deep learning revolution the field of inverse problems has seen a multitude of publications describing how to utilize these new one capabilities to solve inverse problems in general, see, e.g.,~\cite{lunz2018adversarial, adler2018learned, jin2017deep, hammernik2018learning, zhang2017beyond, ye2018deep, arridge2019solving}, and denoising in particular, e.g.,~\cite{zhang2017beyond, kobler2017variational}. And although there has been some progress with unsupervised methods, in particular for denoising~\cite{soltanayev2018training, cha2018neural, metzler2018unsupervised, lehtinen2018noise2noise}, most methods either require some kind of ground truth data or, as already mentioned, make various restrictive assumptions on the type of noise and data they can handle. Most of the general inverse problem methods even require training data of them in the form of $\left(\text{ground truth}, \text{noisy measurement}\right)$-tuples, $(x, y^\delta)$. Since the core of our method rests on the approximation of pushforward operator equations, we consider the most relevant works for this paper to be the original paper on GANs~\cite{goodfellow2014generative}, on WGANs~\cite{arjovsky2017wasserstein}, and the WGAN-GP paper~\cite{gulrajani2017improved}, on which our algorithmic approach is based. We would also like to point out that~\cite{bora2018ambientgan} utilizes an idea similar to our observation 1.

\section{Idea and theory}
\subsection{A reminder about (Wasserstein) GANs}
``Classically'' a GAN~\cite{goodfellow2014generative} (Generative Adversarial Network) is a machine learning setup where two neural networks engage in a minimax game. One of the networks, the generator $G:Y\to Y$, is given random samples from a simple, known distribution $p_\eta$ (e.g., Gaussian) and is supposed to generate a sample from a distribution $p_y$, which one only knows in terms of samples. The other network, the discriminator $C:Y\to\mathbb{R}$ (critic, in the case of a WGAN), has the task to determine whether it sees a sample $G(\eta)$ or a sample $y\sim p_y$. One constructs these networks by pinning them against each other in a minimax game.

More formally, \textbf{the goal of a GAN is} to construct a mapping $G^*:Y\to Y$ with the property that
\begin{equation}
    G^*_\#p_\eta = p_y,
    \label{eq:formalized0}
\end{equation}
where $G^*_\#p_\eta$ denotes the pushforward measure of $p_\eta$ with respect to $G^*$. One does this by minimizing a distance/divergence $D$ between these two distributions, i.e., one tries to solve
\begin{equation}
    \arg\min_G D\left(p_y\right\|G_\#p_\eta).
\end{equation}
The minimax setting emerges since, usually, the computation of the distance/divergence involves a maximization problem.

From now on, we will focus on the setting, where $D$ is the Wasserstein-$1$ distance, known as Wasserstein GAN~\cite{arjovsky2017wasserstein}. WGANs were shown to have favorable properties relative to the ``classical'' GAN setting, which utilizes the Jensen–Shannon divergence~\cite{goodfellow2014generative}. In the WGAN setting the minimax problem  can be derived via the Kantorovich–Rubinstein duality theorem, yielding
\begin{equation}
    \min_G \max_{\substack{\|C_y\|_L\le1}}
    \E_{\substack{y \sim p_y \\
                  \eta \sim p_{\eta}}}
    C_{y}(y) - C_{y}\left(G(\eta)\right).
    \label{eq:wgan}
\end{equation}

\begin{figure}[ht]
    \centering
    \includegraphics[trim={5cm 20.5cm 11cm 4.3cm},clip,width=.3\textwidth]{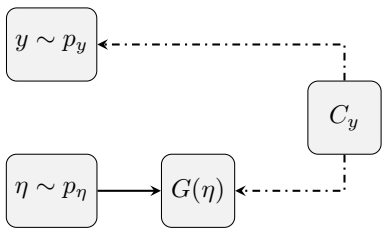}
    \caption{(Wasserstein) GAN training setup.}
    \label{fig:wgan_training_setup_graph}
\end{figure}

We visualize this minimax game, where $C_y$ is a $1$-Lipschitz continuous function that ``tries to find differences between samples from $y \sim p_y$ and $G_\#p_\eta$'' in Figure~\ref{fig:wgan_training_setup_graph}. Another interpretation of Equation~\eqref{eq:wgan} is that WGANs are minimizing the cost of an optimal transport between the distributions $G_\#p_\eta$ and $p_y$.

We will now move on to first formalize our observations about an ideal denoiser $G^*$ from the introduction and then outline how the idea of a GAN -- more specifically, we will use a WGAN -- can be generalized to a setting that allows us to use these observations.

\subsection{Ground truth free denoising via dual critics}
We will now formalize the main idea of the paper. As described in the introduction, our goal is to find a denoiser $G:Y\to Y$ that denoises samples from $y^\delta\sim p_{y^\delta}$, which were created by a measurement process, which added independent noise $\eta\sim p_\eta$ to clean samples $y\sim p_y$. We set out to do so only by using samples from $p_{y^\delta}$ and from $p_\eta$ and -- critically -- no samples from $p_y$.

First, we would like to point out that in this additive noise setup we have
\begin{equation}
    p_{y^\delta} = p_y * p_\eta,
    \label{eq:convolution}
\end{equation}
where ``$*$'' denotes the convolution operator.

Equation~\eqref{eq:convolution} suggests that, as long as we assume that $p_\eta$ has no vanishing frequencies where $p_y$ has non-vanishing frequencies, $p_{y^\delta}$ and $p_\eta$ uniquely determine $p_y$. This follows directly from the convolution theorem~\cite{bracewell2004fourier}. This short discussion demonstrates that it is not unreasonable to train a denoiser solely from noisy and noise samples without any clean samples.

We will now formalize our observations 1 and 2 from the introduction. Observation 0 we already formalized with Equation~\eqref{eq:formalized0} (reading $\eta$ in the current context). \textbf{Observation 1 can be formalised with the expression}
\begin{equation}
\left(G^*_\#p_{y^\delta}\right) * p_\eta  = p_{y^\delta},
\label{eq:observation1}
\end{equation}
i.e., if we ``renoise'' the output of the denoiser, we want to have a noisy sample.
\textbf{Observation 2 can be formalized with the expression}
\begin{equation}
\left(\id - G^*\right)_\#p_{y^\delta} = p_\eta,
\label{eq:observation2}
\end{equation}
i.e., the part that the denoiser removed should look like noise.

Following the WGAN idea, we try to compute a $G$ with the properties 1 and 2 by aiming to minimize the two Wasserstein distances
\begin{equation}
    D\left(p_{y^\delta} \big{\|} \left(G_\#p_{y^\delta}\right) * p_\eta\right)
\end{equation}
and
\begin{equation}
    D\left(p_\eta \big{\|} \left(\id - G\right)_\#p_{y^\delta}\right).
\end{equation}
To do so jointly, \textbf{we choose the loss function of our generator/denoiser to be}
\begin{align}
    \arg\min_G\ &D\left(p_{y^\delta} \big{\|} \left(G_\#p_{y^\delta}\right) * p_\eta\right)\nonumber\\
    + &D\left(p_\eta \big{\|} \left(\id - G\right)_\#p_{y^\delta}\right).
\end{align}

As usual for WGANs, we can use the Kantorovich-Rubinstein duality to rewrite this as the following minimax problem
\begin{align}
\min_G \max_{\substack{\|C_{y^\delta}\|_L\le1,\\ \|C_\eta\|_L\le1}}
&\E_{\substack{y^\delta \sim p_{y^\delta} \\
              \tilde y^\delta \sim p_{y^\delta} \nonumber\\
              \eta \sim p_{\eta}}}
\ C_{y^\delta}(\tilde y^\delta) - C_{y^\delta}\left(G( y^\delta)+\eta\right)\\
+ &\E_{\substack{y^\delta \sim p_{y^\delta}  \\
              \eta \sim p_{\eta}}}
              C_\eta(\eta) - C_\eta\left(y^\delta - G(y^\delta)\right).
\label{eq:loss_function}
\end{align}

\begin{figure}[ht]
    \centering
    \includegraphics[trim={5.2cm 15cm 6cm 4.3cm},clip,width=.47\textwidth]{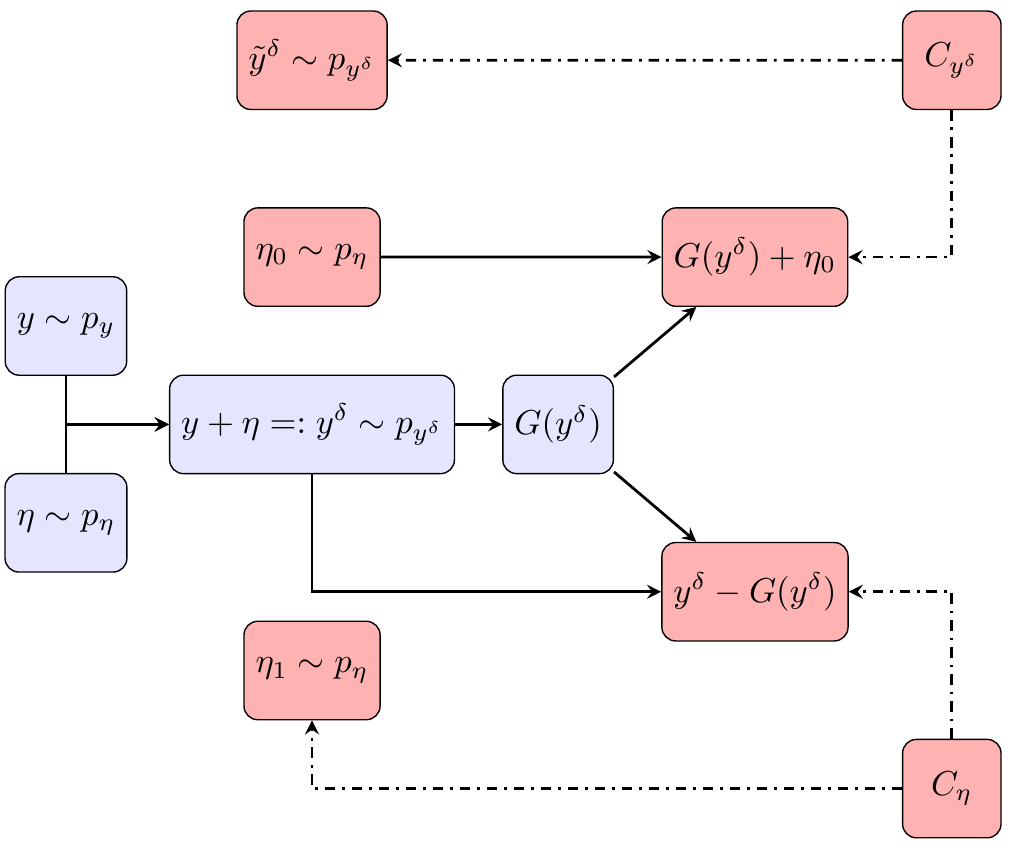}
    \caption{Training setup of the denoiser $G$, as given by Equation~\eqref{eq:loss_function}. Compare with Figure~\ref{fig:wgan_training_setup_graph} for a ``classical'' WGAN. Figure~\ref{fig:example_application_graph} exemplifies the blue part of the graph. The red part is only used during training.}
    \label{fig:training_setup_graph}
\end{figure}

Analogously to the visualization of Equation~\eqref{eq:wgan}, in Figure~\ref{fig:wgan_training_setup_graph}, we visualize Equation~\eqref{eq:loss_function} in Figure~\ref{fig:training_setup_graph}.

Before we move on to numerical experiments, we now want to investigate a simple case analytically.

\subsection{Analysis of a simple case}
To improve our intuition for the denoiser resulting from our two observations, we now want to investigate a simple case analytically. For the analysis, we assume $G:Y\to Y$ and $I-G$ to be linear and positive definite. Further we assume $p_y$ to be $\mathcal{N}(0, I)$ and $p_\eta$ to be $\mathcal{N}(0, \sigma^2I)$.

\begin{figure}
    \centering
    \includegraphics[trim={0cm .3cm 0cm 0cm},clip,width=.45\textwidth]{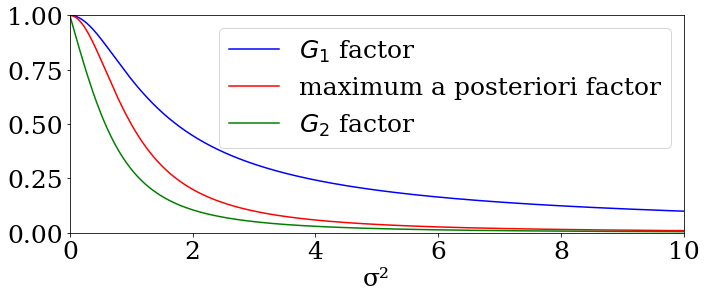}
    \caption{Comparison of the factors of the linear denoisers $G_1$, $G_2$, and the maximum a posteriori one.}
    \label{fig:factors_of_linear_denoisers}
\end{figure}

For this simple case the first observation alone, i.e., Equation~\eqref{eq:observation1}, leads -- via a simple change of variable -- to the denoiser
\begin{equation}
    G_1(y^\delta) = \left(\sqrt{1 + \sigma^2}\right)^{-1} y^\delta,
\end{equation}
whereas the second observation on its own yields the denoiser
\begin{equation}
    G_2(y^\delta) = \left(1 - \sigma/\sqrt{1 + \sigma^2}\right) y^\delta.
\end{equation}
This proves that there is -- somewhat unsurprisingly -- no ideal linear denoising (in the sense that it fulfills both our observations).
Nevertheless we think this gives an insightful perspective on how the solution for a more general model for $G$ might behave; not least because $\frac{1}{\sqrt{1 + \sigma^2}}$ majorizes and $1 - \frac{\sigma}{\sqrt{1 + \sigma^2}} y^\delta$ minorizes $\frac{1}{1+\sigma^2}$, which is the factor given by a maximum a posteriori estimate~\cite{dashti2016bayesian}, which can be derived by minimizing the expression
$\frac{1}{2}\|y-y^\delta\|^2 + \frac{\sigma^2}{2}\|y\|^2.$
For a visualization see Figure~\ref{fig:factors_of_linear_denoisers}.

We will now move on to the algorithmic specificities on how to solve the minimax problem given by Equation~\eqref{eq:loss_function} in the general case, i.e., the training of a denoising network $G$.

\section{Implementation and experiments}
\subsection{Numerical approach}
Before we tackle Equation~\eqref{eq:loss_function} algorithmically, we briefly review the existing techniques for WGANs on which our approach rests. A core problem of the minimax problem in Equation~\eqref{eq:wgan} is that it is a constraint optimization, which requires the critics $C_{y^\delta}$ and $C_\eta$ to be $1$-Lipschitz. Successfully replacing this hard constraint in~\eqref{eq:wgan} by a penalty term, resulting in the unconstrained problem
\begin{align}
    \min_G \max_{\substack{C_y}}
    &\E_{\substack{y \sim p_y \\
                  \eta \sim p_{\eta}}}
    C_{y}(y) - C_{y}\left(G(\eta)\right)\\
    + \lambda_{C_y} &\E_{\tilde y\sim p_{\tilde y}}\left(\|\nabla_{\tilde y}C_y({\tilde y})\| - 1\right)^2,
\end{align}
marks a breakthrough in WGAN research~\cite{gulrajani2017improved}. Here $p_{\tilde y}$ is defined by uniformly sampling along lines between samples of the distributions $p_y$ and $G_\#p_\eta$.

In \cite{petzka2017regularization} the authors show that the training of a WGAN can benefit from convexifying the penalty term, which yields the objective
\begin{align}
    \min_G \max_{\substack{C_y}}
    &\E_{\substack{y \sim p_y \\
                  \eta \sim p_{\eta}}}
    C_{y}(y) - C_{y}\left(G(\eta)\right)\\
    + \lambda_{C_y} &\E_{\tilde y\sim p_{\tilde y}}\relu\left(\|\nabla_{\tilde y}C_y({\tilde y})\| - 1\right)^2.
    \label{eq:wgan_convex}
\end{align}

We use this strategy also for our dual critics approach, which relaxes the Equation~\eqref{eq:loss_function} to the unconstrained problem
\begin{align}
\min_G \max_{\substack{C_{y^\delta}, C_\eta}}
&\E_{\substack{y^\delta \sim p_{y^\delta} \\
              \tilde y^\delta \sim p_{y^\delta} \\
              \eta \sim p_{\eta}}}
\ C_{y^\delta}(\tilde y^\delta) - C_{y^\delta}\left(G( y^\delta)+\eta\right) \nonumber\\
+ &\E_{\substack{y^\delta \sim p_{y^\delta} \\
              \eta \sim p_{\eta}}}
\ C_\eta(\eta) - C_\eta\left(y^\delta - G(y^\delta)\right) \nonumber\\
+ \lambda_{C_{y^\delta}} &\E_{\tilde y^\delta\sim p_{\tilde y^\delta}} \relu\left(\|\nabla_{\tilde y^\delta}C_{y^\delta}({\tilde y^\delta})\| - 1\right)^2 \nonumber\\
+ \lambda_{C_{\eta}} &\E_{\tilde \eta\sim p_{\tilde \eta}} \relu\left(\|\nabla_{\tilde \eta}C_{\eta}({\tilde \eta})\| - 1\right)^2,
\label{eq:loss_function_relaxed}
\end{align}
where the definition of $p_{\tilde y}$ in Equation~\eqref{eq:wgan_convex} applies mutatis mutandis to $p_{\tilde y^\delta}$ and $p_{\tilde \eta}$. In practice, for simplicity, we will set $\lambda_{C_{y^\delta}}=\lambda_{C_\eta}$ and denote the variable by $\lambda$. Overall this yields Algorithm~\ref{alg:dual_critics}, which is a modification of the WGAN-GP algorithm presented in~\cite{gulrajani2017improved}.

\begin{algorithm}[tb]
\caption{WGAN algorithm to tackle Equation~\eqref{eq:loss_function}.}
\begin{algorithmic}
\REQUIRE Samples from $p_{y^\delta}$, samples from $p_\eta$, \\$\gamma_{\mbox{Adam}_*} = (\alpha, \beta_1, \beta_2) = (2\cdot10^{-4}, .5, .9)$, \\$k=\mbox{batch\_size}$, $\lambda=10$, \\ initializations of the model parameters $\theta_G$, $\theta_{C_{\eta}}$, $\theta_{C_{y^\delta}}$
\WHILE{not converged}
    \FOR{$i = 1, \dots, k$}
        \STATE $y_0^\delta \sim p_{y^\delta}$,\ \
               $\eta_0 \sim p_\eta$,\ \ $\epsilon \sim U[0,1]$
        \STATE $\hat y = G(y^\delta)$
        \STATE // Compute loss for $C_{y^\delta}$.
        \STATE $\overline y^\delta \leftarrow \epsilon y_0^\delta + (1 - \epsilon)(\hat y + \eta_0)$
        \STATE $L_{C_{y^\delta}} \leftarrow \relu\left(\|\nabla_{\overline y^\delta}C_{y^\delta}(\overline y^\delta)\| - 1\right)^2$
        \STATE $L_{C_{y^\delta}}^{(i)} \leftarrow C_{y^\delta}(\hat y + \eta_0) - C_{y^\delta}(y^\delta) + \lambda L_{C_{y^\delta}}$
        \STATE // Compute loss for $C_\eta$.
        \STATE $\overline\eta \leftarrow \epsilon \eta_0 + (1 - \epsilon)(y^\delta - \hat y^\delta)$
        \STATE $L_{C_\eta} \leftarrow \relu\left(\|\nabla_{\overline \eta}C_\eta(\overline \eta)\| - 1\right)^2$
        \STATE $L_{C_\eta}^{(i)} \leftarrow C_\eta(y^\delta - \hat y^\delta) - C_\eta(\eta_0) + \lambda L_{C_\eta}$
    \ENDFOR
    \STATE $\theta_{C_{y^\delta}} \leftarrow \mbox{Adam}_{C_{y^\delta}}\left(\nabla_{\theta_{C_{y^\delta}}}\frac{1}{k}\sum_i L_{C_{y^\delta}}^{(i)}, \theta_{C_{y^\delta}}, \gamma\right)$
    \STATE $\theta_{C_\eta} \leftarrow \mbox{Adam}_{C_\eta}\left(\nabla_{\theta_{C_\eta}}\frac{1}{k}\sum_i L_{C_\eta}^{(i)}, \theta_{C_\eta}, \gamma\right)$
    \STATE // Compute loss for $G$.
    \FOR{$i = 1, \dots, k$}
        \STATE $y_1^\delta \sim p_{y^\delta}$,\ \ $\eta_1 \sim p_\eta$
        \STATE $\hat y \leftarrow G(y_1^\delta)$
        \STATE $L_G^{(i)} \leftarrow -C_{y^\delta}(\hat y + \eta_1) - C_\eta(y_1^\delta - \hat y)$
    \ENDFOR
    \STATE $\theta_G \leftarrow \mbox{Adam}_G\left(\nabla_{\theta_G}\frac{1}{k}\sum_i L_G^{(i)}, \theta_G, \gamma\right)$
\ENDWHILE
\end{algorithmic}
\label{alg:dual_critics}
\end{algorithm}

In the next subsection, we will describe general details about our implementation of the algorithm and the following experiments.

\subsection{General implementation details}
We will now document general details about all the following numerical experiments. Each of our experiments was run on an Nvidia GeForce GTX 1080 Ti and used the Adam optimizer~\cite{kingma2014adam} with the settings as outlined in Algorithm~\ref{alg:dual_critics}.

For simplicity, all critics in our experiments are based on the same critic architecture (up to the dimensionality of the convolutions). The architecture is a convolutional ResNet~\cite{he2016deep} with a final scalar-valued dense linear layer. All convolutional kernels used are of size 3. The network starts with a linear convolutional layer increasing the number of $1$ or $3$ channels to $16$. $5$ ResBlocks follow it, each based on two internal convolutions followed by layer-normalizations~\cite{ba2016layer} with the first layer-normalization itself being followed by a $\leakyrelu$~\cite{xu2015empirical} activation. The ResBlock is concluded by a subsampling of factor $2$ and a doubling of the channels via ``self-concatenation''. For a visualization, parameter settings and training details see Figure~\ref{fig:critic_architecture}, Figure~\ref{fig:resBlock}, and Table~\ref{tab:parameters} in the appendix.

We also compare all our networks with a -- via an $\ell_2$ loss -- supervised-trained version, and a SURE estimator trained version~\cite{metzler2018unsupervised} of the respective generator network. Further, the standard deviation for BM3D~\cite{dabov2007image} and the SURE training are computed exactly for the type of noise used. The total variation~\cite{rudin1992nonlinear}) penalty parameter will be optimized via an exponential line search to maximize the PSNR (peak signal-to-noise ratio) with respect to the clean $y$ and (the later introduced) $x$ respectively.

Now, in all the following subsections of this section, we will describe specific experiments and results, beginning with a simple one-dimensional example, followed by image denoising and image deblurring.

\subsection{Denoising of periodic one-dimensional signals}
\label{subsection:1d}
We will now present our first numerical experiment. The task is a denoising task; more specifically, we aim at removing additive independent Gaussian noise $\mathcal{N}(0, I)$ from functions $\sin(2\pi\nu \cdot)$, for which we sample $\nu$ uniformly over the interval $[0, 5]$. In practice, we realize this function by a vector of length $128$, which represents equidistant samples of the function on the interval $[0, 1]$. We train in the setting described above and, for simplicity, use an autoencoder architecture~\cite{baldi2012autoencoders} as the generator.

The auto-encoder architecture, see Figure~\ref{fig:1dim_generator_architecture} in the appendix for more details, starts with three convolutional blocks, which lead to a bottleneck, which consist of two dense $\relu$-layers. It then expands the bottleneck again, via another three convolutional blocks, back to the original input shape. In this setting, each convolutional block consists out of a one-dimensional convolutional layer with kernel size $3$ (except the last one for which it is $1$) followed by a layer-normalization and a $\relu$-activation.

The results of the experiment are displayed in Table~\ref{tab:results_1d}. As we see, the method gives us a significant improvement in the PSNR. It improves the PSNR of the noisy measurements, with an average PSNR of $6.1$, to an average PSNR of $25.3$ and (probably due to the simplicity of the data) even outperforms the for Gaussian noise specialized SURE estimator.

\begin{table*}%[t]
    \centering
    \caption{Reconstructions of sines. (Stats over 10,000 samples.) Here, we compare (from left to right) the ground truth, the noisy measurement, our setup, a supervised setup, and a SURE setup.}
    \begin{tabular}{ccccc}
        $y$ & $y^\delta = y + \eta$ & $G(y^\delta)$ & $G_{\text{supervised}}(y^\delta)$ & $G_{\text{SURE}}(y^\delta)$ \\
        \hline
        \includegraphics[trim={6.88cm 0.42cm 47.04cm 0.3cm},clip, width=.175\textwidth]{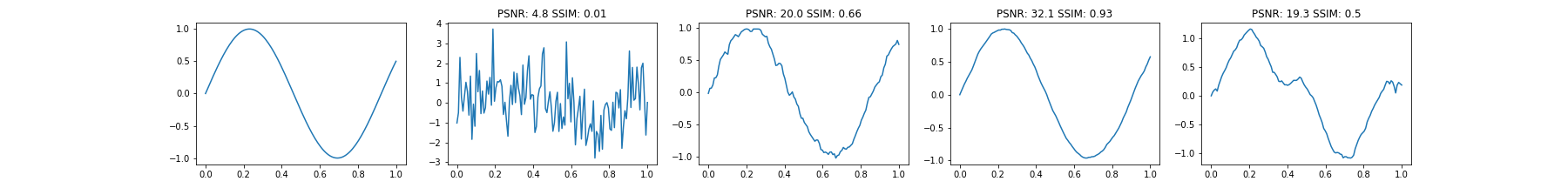} &
        \includegraphics[trim={17.08cm 0.42cm 36.87cm 0.3cm},clip, width=.175\textwidth]{figures/toy_example/example0.png} &
        \includegraphics[trim={27.13cm 0.42cm 26.68cm 0.3cm},clip, width=.175\textwidth]{figures/toy_example/example0.png} &
        \includegraphics[trim={37.33cm 0.42cm 16.51cm 0.3cm},clip, width=.175\textwidth]{figures/toy_example/example0.png} &
        \includegraphics[trim={47.49cm 0.42cm 6.32cm 0.3cm},clip, width=.175\textwidth]{figures/toy_example/example0.png} \\

        \includegraphics[trim={6.88cm 0.42cm 47.04cm 0.3cm},clip, width=.175\textwidth]{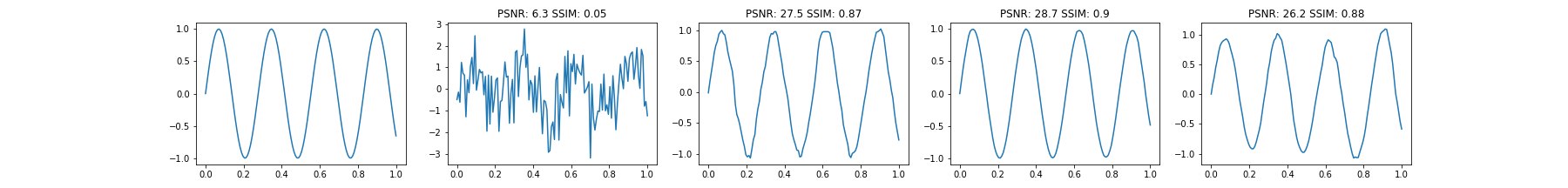} &
        \includegraphics[trim={17.08cm 0.42cm 36.87cm 0.3cm},clip, width=.175\textwidth]{figures/toy_example/example1.png} &
        \includegraphics[trim={27.13cm 0.42cm 26.68cm 0.3cm},clip, width=.175\textwidth]{figures/toy_example/example1.png} &
        \includegraphics[trim={37.33cm 0.42cm 16.51cm 0.3cm},clip, width=.175\textwidth]{figures/toy_example/example1.png} &
        \includegraphics[trim={47.49cm 0.42cm 6.32cm 0.3cm},clip, width=.175\textwidth]{figures/toy_example/example1.png} \\

        \includegraphics[trim={6.88cm 0.42cm 47.04cm 0.3cm},clip, width=.175\textwidth]{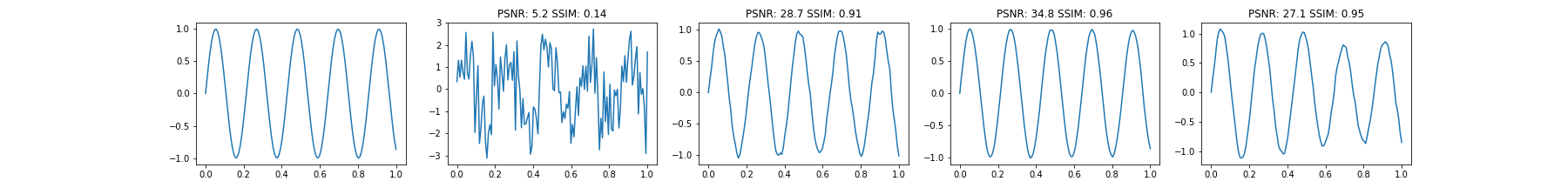} &
        \includegraphics[trim={17.08cm 0.42cm 36.87cm 0.3cm},clip, width=.175\textwidth]{figures/toy_example/example2.png} &
        \includegraphics[trim={27.13cm 0.42cm 26.68cm 0.3cm},clip, width=.175\textwidth]{figures/toy_example/example2.png} &
        \includegraphics[trim={37.33cm 0.42cm 16.51cm 0.3cm},clip, width=.175\textwidth]{figures/toy_example/example2.png} &
        \includegraphics[trim={47.49cm 0.42cm 6.32cm 0.3cm},clip, width=.175\textwidth]{figures/toy_example/example2.png} \\

        \hline
        \multirow{2}{*}{\textbf{PSNR stats}:}
        & \ \ mean:     $ 6.1$ & \ \ mean:     $25.3$ & \ \ mean:     $31.0$ & \ \ mean:     $23.4$ \\
        & median:   $ 6.0$ & median:   $25.5$ & median:   $30.3$ & median:   $23.4$\\
         \hline
        \multirow{2}{*}{\textbf{SSIM stats}:}
        & \ \ mean:   $0.03$ & \ \ mean:     $0.79$ & \ \ mean:     $0.90$ & \ \ mean:     $0.78$ \\
        & median: $0.03$ & median:   $0.81$ & median:   $0.93$ & median:   $0.80$ \\
    \end{tabular}
    \label{tab:results_1d}
\end{table*}

\subsection{Denoising of images}
\label{subsection:denoise_images}
In this subsection, we will present our results for image denoising using Algorithm~\ref{alg:dual_critics}. We present results for Gaussian and spatially dependent non-Gaussian noise. As a generator we will use a simple UNet-like~\cite{ronneberger2015u} architecture, for more details see Figure~\ref{fig:image_generator_architecture} in the appendix.
We use the image data set STL-10~\cite{coates2011analysis} and train for $2$ different noise settings.

The first noise we considered (and train a denoising network for) is Gaussian noise with a standard deviation of $0.08$, which is firmly in the comfort zone of classical denoisers like BM3D~\cite{dabov2007image}. We present the results on this in Table~\ref{tab:results_I_Gaussian}. As the figures show, the approach performs well but is slightly worse than BM3D and SURE, which are specialized for Gaussian noise.

The next noise we considered we call ``mixed noise.'' It is Gaussian noise with a standard deviation that is uniformly sampled from the interval $[0, 0.2]$ plus spatially dependent noise, which we will refer to as ``localized noise''. We create the ``localized noise'' for each RGB color channel separately. We do so by uniformly choosing a random position in the image as the mean of a Gaussian with a standard deviation of $5$. We then sample $500$ spatial coordinates for each channel from this Gaussian (if necessary, project them back into the image domain) and add at these positions Gaussian noise with a standard deviation of $0.5$. We devised this type of noise for mainly two reasons: First, we want to demonstrate that our algorithm can deal with spatially dependent noise and second, it would be quite hard to come up with a proper denoiser for this by hand, whereas learning ``only'' requires the appropriate training data. We present the results on this in Table~\ref{tab:results_I_Mixed}. While comparing with the other unsupervised techniques might not be entirely fair, since they (like most others) are specialized for a specific type of noise, the figures show that here our approach outperforms all other unsupervised methods by a significant margin.

As a third comparison, we use the networks trained on the mixed noise and apply them to the localized noise only (i.e., without the Gaussian noise with uniformly sampled standard deviation). The results can be seen in Table~\ref{tab:results_I_localized_noise}. As displayed, in this setting, our approach not only outperforms the other unsupervised approaches by a wide margin but even the supervised comparison, which was trained with ground truth and an, admittedly for non-Gaussian noise, sub-optimal, $\ell_2$ loss.

\begin{table*}
    \centering
    \caption{Reconstructions of Gaussian-noisy images. (Stats over 256 samples.) Here, we compare (from left to right) the ground truth, the noisy measurement, our setup, BM3D, the median filter, a supervised setup, and a SURE setup.}
    \begin{tabular}{ccccccc}
        $y$ & $y^\delta = y + \eta$ & $G(y^\delta)$ & $\mbox{BM3D}(y^\delta, \sigma)$ & $\mbox{MEDIAN}(y^\delta)$ & $G_{\text{supervised}}(y^\delta)$ & $G_{\text{SURE}}(y^\delta, \sigma)$  \\
        \hline\\ [-2ex]

        \includegraphics[trim={9.53cm 4.06cm 59.46cm 3.4cm},clip, width=.12\textwidth]{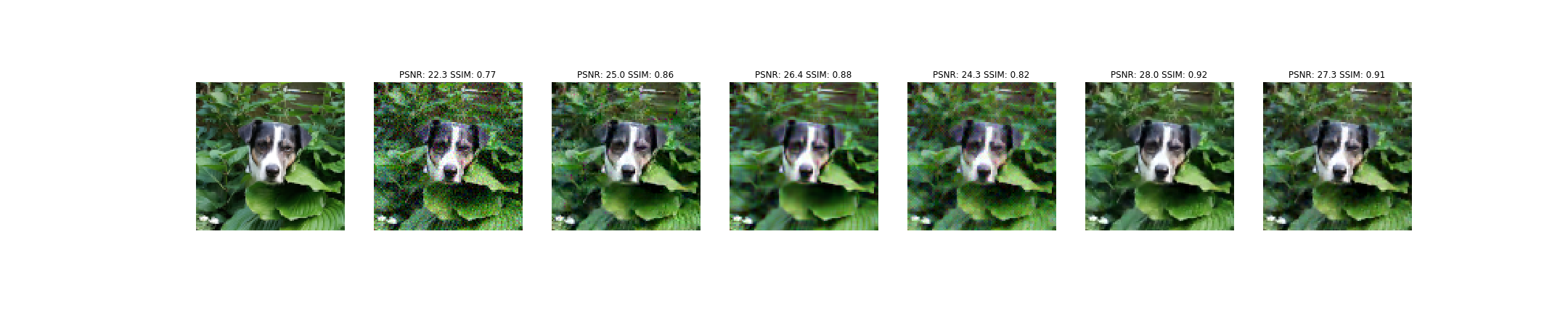} &
        \includegraphics[trim={18.17cm 4.06cm 50.81cm 3.4cm},clip, width=.12\textwidth]{figures/photos/I_Gaussian0.png} &
        \includegraphics[trim={26.82cm 4.06cm 42.17cm 3.4cm},clip, width=.12\textwidth]{figures/photos/I_Gaussian0.png} &
        \includegraphics[trim={35.46cm 4.06cm 33.52cm 3.4cm},clip, width=.12\textwidth]{figures/photos/I_Gaussian0.png} &
        \includegraphics[trim={44.11cm 4.06cm 24.88cm 3.4cm},clip, width=.12\textwidth]{figures/photos/I_Gaussian0.png} &
        \includegraphics[trim={52.75cm 4.06cm 16.23cm 3.4cm},clip, width=.12\textwidth]{figures/photos/I_Gaussian0.png} &
        \includegraphics[trim={61.4cm 4.06cm 7.59cm 3.4cm},clip, width=.12\textwidth]{figures/photos/I_Gaussian0.png} \\

        \includegraphics[trim={9.53cm 4.06cm 59.46cm 3.4cm},clip, width=.12\textwidth]{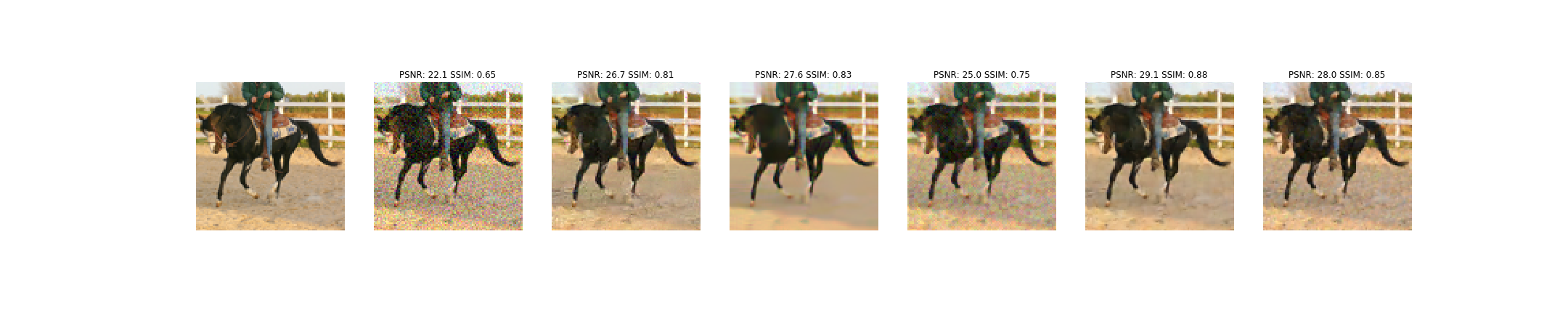} &
        \includegraphics[trim={18.17cm 4.06cm 50.81cm 3.4cm},clip, width=.12\textwidth]{figures/photos/I_Gaussian1.png} &
        \includegraphics[trim={26.82cm 4.06cm 42.17cm 3.4cm},clip, width=.12\textwidth]{figures/photos/I_Gaussian1.png} &
        \includegraphics[trim={35.46cm 4.06cm 33.52cm 3.4cm},clip, width=.12\textwidth]{figures/photos/I_Gaussian1.png} &
        \includegraphics[trim={44.11cm 4.06cm 24.88cm 3.4cm},clip, width=.12\textwidth]{figures/photos/I_Gaussian1.png} &
        \includegraphics[trim={52.75cm 4.06cm 16.23cm 3.4cm},clip, width=.12\textwidth]{figures/photos/I_Gaussian1.png} &
        \includegraphics[trim={61.4cm 4.06cm 7.59cm 3.4cm},clip, width=.12\textwidth]{figures/photos/I_Gaussian1.png} \\

        \includegraphics[trim={9.53cm 4.06cm 59.46cm 3.4cm},clip, width=.12\textwidth]{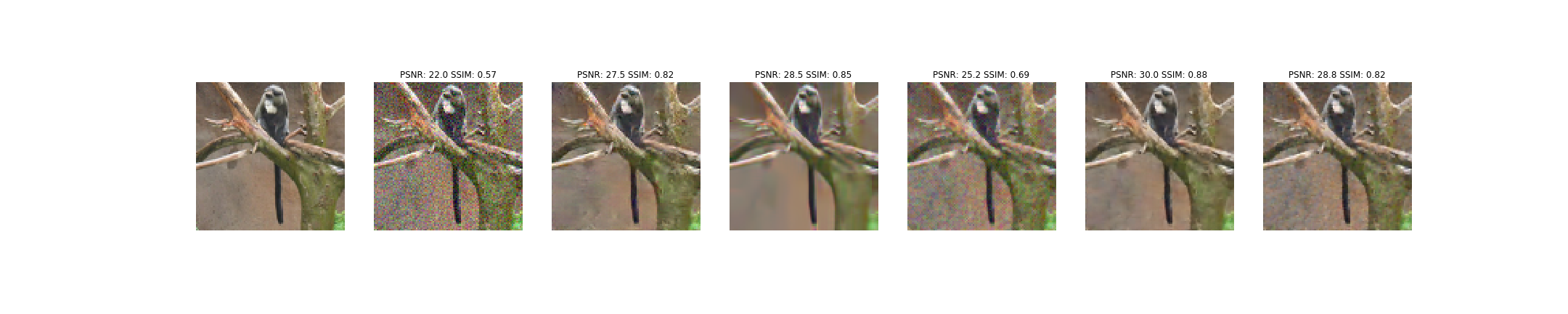} &
        \includegraphics[trim={18.17cm 4.06cm 50.81cm 3.4cm},clip, width=.12\textwidth]{figures/photos/I_Gaussian2.png} &
        \includegraphics[trim={26.82cm 4.06cm 42.17cm 3.4cm},clip, width=.12\textwidth]{figures/photos/I_Gaussian2.png} &
        \includegraphics[trim={35.46cm 4.06cm 33.52cm 3.4cm},clip, width=.12\textwidth]{figures/photos/I_Gaussian2.png} &
        \includegraphics[trim={44.11cm 4.06cm 24.88cm 3.4cm},clip, width=.12\textwidth]{figures/photos/I_Gaussian2.png} &
        \includegraphics[trim={52.75cm 4.06cm 16.23cm 3.4cm},clip, width=.12\textwidth]{figures/photos/I_Gaussian2.png} &
        \includegraphics[trim={61.4cm 4.06cm 7.59cm 3.4cm},clip, width=.12\textwidth]{figures/photos/I_Gaussian2.png} \\

        \hline
        \multirow{2}{*}{\textbf{PSNR stats}:}
        & \ \ mean:     $22.3$& \ \ mean:     $27.7$& \ \ mean:     $28.7$& \ \ mean:     $25.1$& \ \ mean:     $30.1$& \ \ mean:     $28.6$  \\
        & median:     $22.1$& median:     $27.6$& median:     $28.5$& median:     $25.3$& median:     $30.0$& median:     $28.6$  \\
         \hline
        \multirow{2}{*}{\textbf{SSIM stats}:}
        & \ \ mean:     $0.6$& \ \ mean:     $0.85$& \ \ mean:     $0.87$& \ \ mean:     $0.71$& \ \ mean:     $0.9$& \ \ mean:     $0.85$  \\
        & median:     $0.61$& median:     $0.85$& median:     $0.88$& median:     $0.72$& median:     $0.9$& median:     $0.86$  \\
    \end{tabular}
    \label{tab:results_I_Gaussian}
\end{table*}

\begin{table*}
    \centering
    \caption{Reconstructions of Mixed-noisy images. (Stats over 256 samples.)}
    \begin{tabular}{ccccccc}
        $y$ & $y^\delta = y + \eta$ & $G(y^\delta)$ & $\mbox{BM3D}(y^\delta, \sigma)$ & $\mbox{MEDIAN}(y^\delta)$ & $G_{\text{supervised}}(y^\delta)$ & $G_{\text{SURE}}(y^\delta, \sigma)$  \\
        \hline\\ [-2ex]

        \includegraphics[trim={9.53cm 4.06cm 59.46cm 3.4cm},clip, width=.12\textwidth]{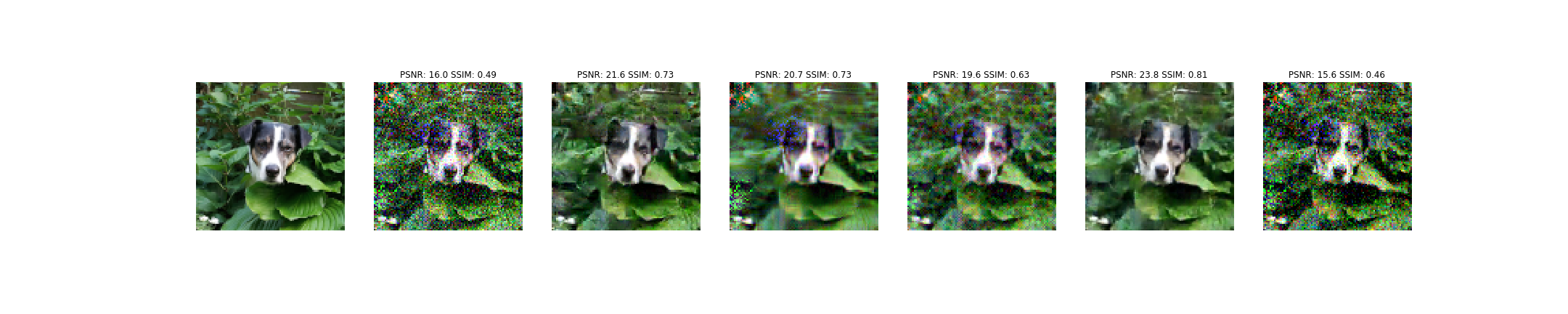} &
        \includegraphics[trim={18.17cm 4.06cm 50.81cm 3.4cm},clip, width=.12\textwidth]{figures/photos/I_Mixed0.png} &
        \includegraphics[trim={26.82cm 4.06cm 42.17cm 3.4cm},clip, width=.12\textwidth]{figures/photos/I_Mixed0.png} &
        \includegraphics[trim={35.46cm 4.06cm 33.52cm 3.4cm},clip, width=.12\textwidth]{figures/photos/I_Mixed0.png} &
        \includegraphics[trim={44.11cm 4.06cm 24.88cm 3.4cm},clip, width=.12\textwidth]{figures/photos/I_Mixed0.png} &
        \includegraphics[trim={52.75cm 4.06cm 16.23cm 3.4cm},clip, width=.12\textwidth]{figures/photos/I_Mixed0.png} &
        \includegraphics[trim={61.4cm 4.06cm 7.59cm 3.4cm},clip, width=.12\textwidth]{figures/photos/I_Mixed0.png} \\

        \includegraphics[trim={9.53cm 4.06cm 59.46cm 3.4cm},clip, width=.12\textwidth]{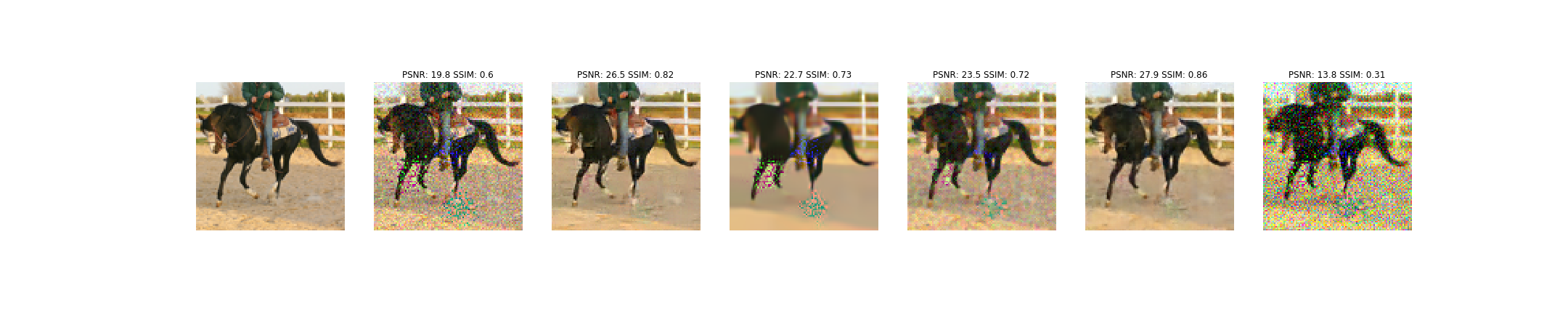} &
        \includegraphics[trim={18.17cm 4.06cm 50.81cm 3.4cm},clip, width=.12\textwidth]{figures/photos/I_Mixed1.png} &
        \includegraphics[trim={26.82cm 4.06cm 42.17cm 3.4cm},clip, width=.12\textwidth]{figures/photos/I_Mixed1.png} &
        \includegraphics[trim={35.46cm 4.06cm 33.52cm 3.4cm},clip, width=.12\textwidth]{figures/photos/I_Mixed1.png} &
        \includegraphics[trim={44.11cm 4.06cm 24.88cm 3.4cm},clip, width=.12\textwidth]{figures/photos/I_Mixed1.png} &
        \includegraphics[trim={52.75cm 4.06cm 16.23cm 3.4cm},clip, width=.12\textwidth]{figures/photos/I_Mixed1.png} &
        \includegraphics[trim={61.4cm 4.06cm 7.59cm 3.4cm},clip, width=.12\textwidth]{figures/photos/I_Mixed1.png} \\

        \includegraphics[trim={9.53cm 4.06cm 59.46cm 3.4cm},clip, width=.12\textwidth]{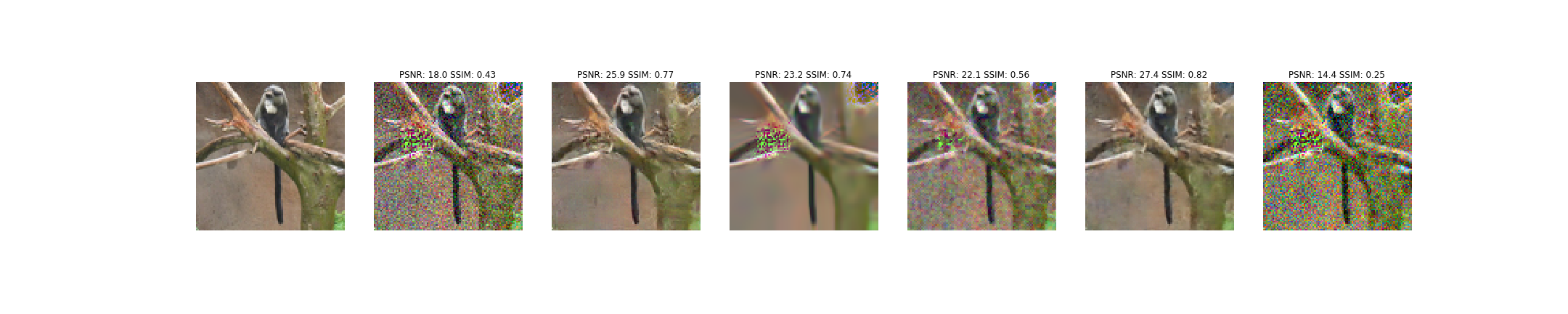} &
        \includegraphics[trim={18.17cm 4.06cm 50.81cm 3.4cm},clip, width=.12\textwidth]{figures/photos/I_Mixed2.png} &
        \includegraphics[trim={26.82cm 4.06cm 42.17cm 3.4cm},clip, width=.12\textwidth]{figures/photos/I_Mixed2.png} &
        \includegraphics[trim={35.46cm 4.06cm 33.52cm 3.4cm},clip, width=.12\textwidth]{figures/photos/I_Mixed2.png} &
        \includegraphics[trim={44.11cm 4.06cm 24.88cm 3.4cm},clip, width=.12\textwidth]{figures/photos/I_Mixed2.png} &
        \includegraphics[trim={52.75cm 4.06cm 16.23cm 3.4cm},clip, width=.12\textwidth]{figures/photos/I_Mixed2.png} &
        \includegraphics[trim={61.4cm 4.06cm 7.59cm 3.4cm},clip, width=.12\textwidth]{figures/photos/I_Mixed2.png} \\

        \hline
        \multirow{2}{*}{\textbf{PSNR stats}:}
        & \ \ mean:     $19.5$& \ \ mean:     $27.5$& \ \ mean:     $22.6$& \ \ mean:     $23.1$& \ \ mean:     $28.5$& \ \ mean:     $13.1$  \\
        & median:     $18.9$& median:     $26.8$& median:     $22.8$& median:     $22.7$& median:     $28.1$& median:     $14.1$  \\
         \hline
        \multirow{2}{*}{\textbf{SSIM stats}:}
        & \ \ mean:     $0.55$& \ \ mean:     $0.82$& \ \ mean:     $0.73$& \ \ mean:     $0.66$& \ \ mean:     $0.85$& \ \ mean:     $0.23$  \\
        & median:     $0.51$& median:     $0.83$& median:     $0.75$& median:     $0.65$& median:     $0.86$& median:     $0.24$  \\
    \end{tabular}
    \label{tab:results_I_Mixed}
\end{table*}

\begin{table*}
    \centering
    \caption{Reconstructions of locally-noisy images. (Stats over 256 samples.)}
    \begin{tabular}{ccccccc}
        $y$ & $y^\delta = y + \eta$ & $G(y^\delta)$ & $\mbox{BM3D}(y^\delta, \sigma)$ & $\mbox{MEDIAN}(y^\delta)$ & $G_{\text{supervised}}(y^\delta)$ & $G_{\text{SURE}}(y^\delta, \sigma)$  \\
        \hline\\ [-2ex]

        \includegraphics[trim={9.53cm 4.06cm 59.46cm 3.4cm},clip, width=.12\textwidth]{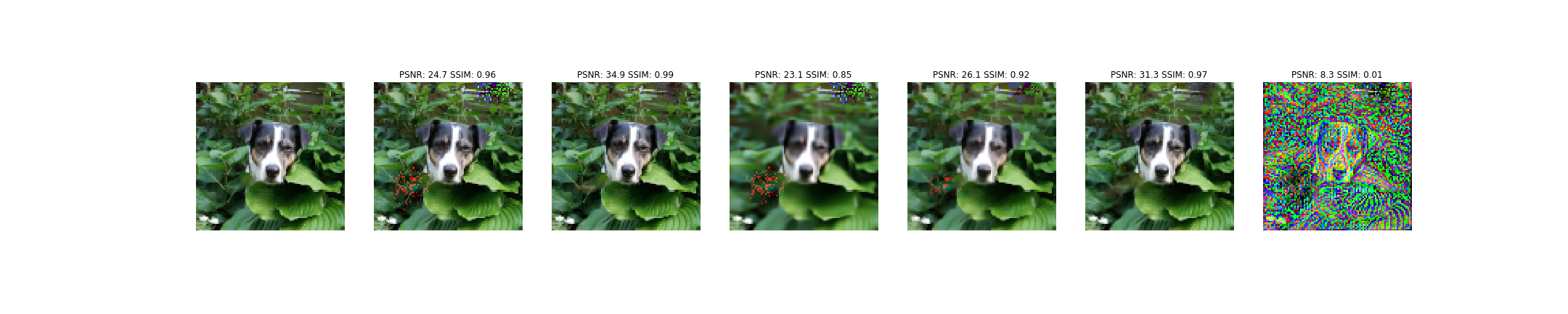} &
        \includegraphics[trim={18.17cm 4.06cm 50.81cm 3.4cm},clip, width=.12\textwidth]{figures/photos/I_Blob0.png} &
        \includegraphics[trim={26.82cm 4.06cm 42.17cm 3.4cm},clip, width=.12\textwidth]{figures/photos/I_Blob0.png} &
        \includegraphics[trim={35.46cm 4.06cm 33.52cm 3.4cm},clip, width=.12\textwidth]{figures/photos/I_Blob0.png} &
        \includegraphics[trim={44.11cm 4.06cm 24.88cm 3.4cm},clip, width=.12\textwidth]{figures/photos/I_Blob0.png} &
        \includegraphics[trim={52.75cm 4.06cm 16.23cm 3.4cm},clip, width=.12\textwidth]{figures/photos/I_Blob0.png} &
        \includegraphics[trim={61.4cm 4.06cm 7.59cm 3.4cm},clip, width=.12\textwidth]{figures/photos/I_Blob0.png} \\

        \includegraphics[trim={9.53cm 4.06cm 59.46cm 3.4cm},clip, width=.12\textwidth]{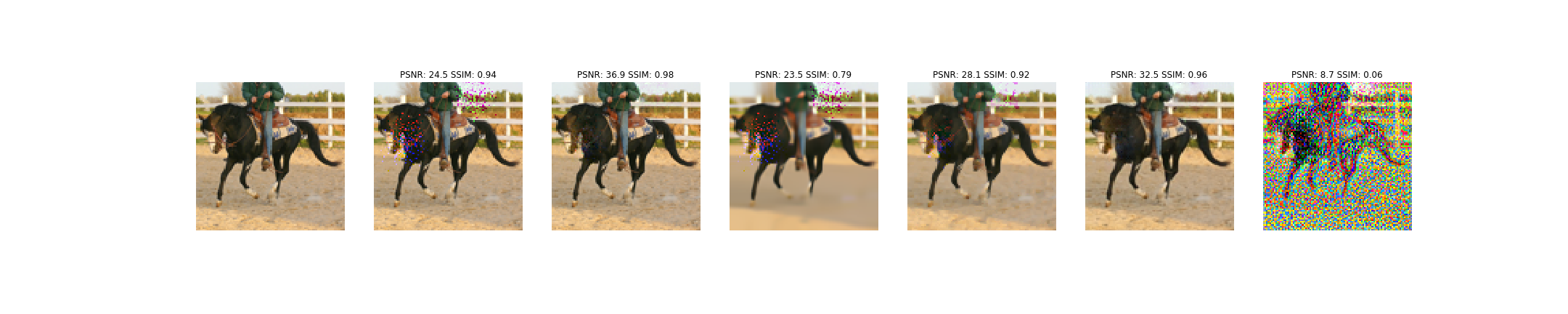} &
        \includegraphics[trim={18.17cm 4.06cm 50.81cm 3.4cm},clip, width=.12\textwidth]{figures/photos/I_Blob1.png} &
        \includegraphics[trim={26.82cm 4.06cm 42.17cm 3.4cm},clip, width=.12\textwidth]{figures/photos/I_Blob1.png} &
        \includegraphics[trim={35.46cm 4.06cm 33.52cm 3.4cm},clip, width=.12\textwidth]{figures/photos/I_Blob1.png} &
        \includegraphics[trim={44.11cm 4.06cm 24.88cm 3.4cm},clip, width=.12\textwidth]{figures/photos/I_Blob1.png} &
        \includegraphics[trim={52.75cm 4.06cm 16.23cm 3.4cm},clip, width=.12\textwidth]{figures/photos/I_Blob1.png} &
        \includegraphics[trim={61.4cm 4.06cm 7.59cm 3.4cm},clip, width=.12\textwidth]{figures/photos/I_Blob1.png} \\

        \includegraphics[trim={9.53cm 4.06cm 59.46cm 3.4cm},clip, width=.12\textwidth]{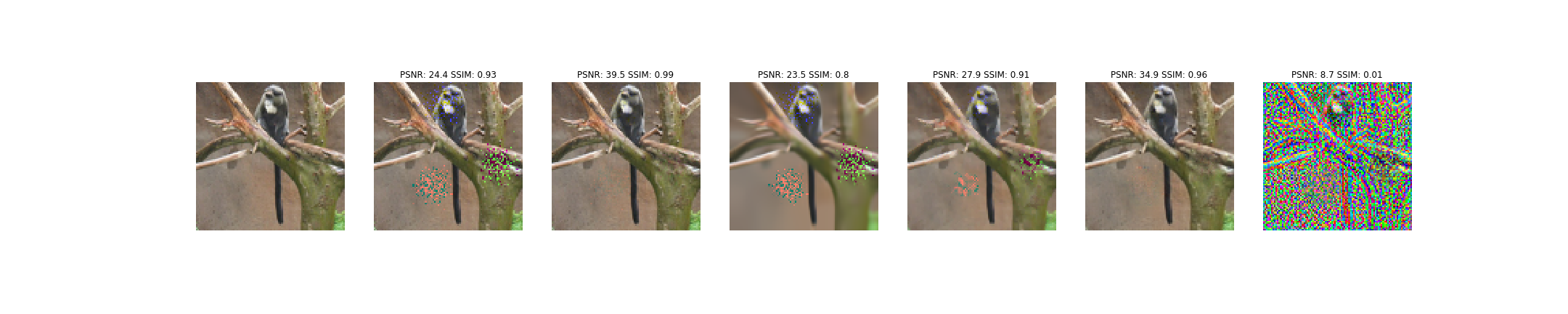} &
        \includegraphics[trim={18.17cm 4.06cm 50.81cm 3.4cm},clip, width=.12\textwidth]{figures/photos/I_Blob2.png} &
        \includegraphics[trim={26.82cm 4.06cm 42.17cm 3.4cm},clip, width=.12\textwidth]{figures/photos/I_Blob2.png} &
        \includegraphics[trim={35.46cm 4.06cm 33.52cm 3.4cm},clip, width=.12\textwidth]{figures/photos/I_Blob2.png} &
        \includegraphics[trim={44.11cm 4.06cm 24.88cm 3.4cm},clip, width=.12\textwidth]{figures/photos/I_Blob2.png} &
        \includegraphics[trim={52.75cm 4.06cm 16.23cm 3.4cm},clip, width=.12\textwidth]{figures/photos/I_Blob2.png} &
        \includegraphics[trim={61.4cm 4.06cm 7.59cm 3.4cm},clip, width=.12\textwidth]{figures/photos/I_Blob2.png} \\

        \hline
        \multirow{2}{*}{\textbf{PSNR stats}:}
        & \ \ mean:     $24.8$& \ \ mean:     $38.3$& \ \ mean:     $23.8$& \ \ mean:     $27.6$& \ \ mean:     $33.8$& \ \ mean:     $8.4$  \\
        & median:     $24.7$& median:     $38.4$& median:     $23.8$& median:     $27.8$& median:     $34.0$& median:     $8.4$  \\
         \hline
        \multirow{2}{*}{\textbf{SSIM stats}:}
        & \ \ mean:     $0.95$& \ \ mean:     $0.98$& \ \ mean:     $0.83$& \ \ mean:     $0.92$& \ \ mean:     $0.96$& \ \ mean:     $0.03$  \\
        & median:     $0.95$& median:     $0.99$& median:     $0.85$& median:     $0.93$& median:     $0.97$& median:     $0.02$  \\
    \end{tabular}
    \label{tab:results_I_localized_noise}
\end{table*}

\subsection{Denoising of measurements} 
One advantage of our denoising method is that it can be applied to any kind of data and additive noise, which makes it useful for all kinds of measurement data in signal processing (as in Subsection~\ref{subsection:1d}) and inverse problems. Both of these domains often deal with highly task-specific measurements and noise for which usually neither a handcrafted denoiser nor ground truth data exists.

Therefore, we will use this subsection to follow-up on the idea of using a wavelet–vaguelette decomposition~\cite{donoho1995nonlinear} to denoise measurement data prior to reconstruction, which was generalized and theoretically analyzed in~\cite{klann2006two}. Specifically, we will consider the following variational formulation:
\begin{equation}
    \mbox{R}_\lambda(G_\delta(y^\delta)) := \arg\min_x \frac{1}{2}\|Ax-G_\delta(y^\delta)\|^2 + \lambda J(x)
    \label{eq:denoising_reconstruction}
\end{equation}
Here $A:X\to Y$ is a (linear) forward/measurement operator, $G_\delta$ a noise level dependent denoiser and $J:X\to\mathbb{R}$ some regularizing convex functional. The goal is to find an $x$ that ``plausibly explains'' the noisy measurement $y^\delta = y + \eta = Ax + \eta$, where $\eta$ some additive noise. In the following we will always use
$J(x) := \mbox{TV}(x) := \|\nabla x\|_1,$ i.e., regularization of $x$ by total variation~\cite{rudin1992nonlinear} and $A:X\to X$ the blurring operator with the uniform $3\times3$ kernel. Note, that here the denoiser plays the role of a preprocessing step.

As in Subsection~\ref{subsection:denoise_images}, we will use STL-10 and the three different kinds of noise, i.e., Gaussian, mixed, and localized noise, for which we only train networks for the first two and apply the one for mixed noise also to the localized one. Here, in contrast to Subsection~\ref{subsection:denoise_images}, the $y$'s are not given as samples from STL-10, but blurry versions of them, given via $Ax$, where the $x$ represent the clean image samples. We therefore deal with noisy, blurry measurements $y^\delta = Ax + \eta$ which we first denoise and then apply a TV reconstruction to, see Equation~\eqref{eq:denoising_reconstruction}.
We present the results for the Gaussian denoising in Table~\ref{tab:results_3x3_Gaussian}. As visible here, denoising of any kind does not make much of a difference, which is unsurprising, since our fidelity term in Equation~\eqref{eq:denoising_reconstruction} is $\ell_2$, which is optimal for Gaussian noise. The results start to look differently for the mixed noise in Table~\ref{tab:results_3x3_Mixed}. This improvement is also unsurprising, since here neither $\ell_2$ is the optimal fidelity term nor -- even within this case explicit knowledge of the noise distribution -- is it trivial to write down an optimal fidelity term. Unsurprisingly, the advantage becomes even more prominent for the purely localized noise, as displayed in Table~\ref{tab:results_3x3_localized_noise}.

We will now outline how this general approach can be adapted to not only apply to additive but also multiplicative noise.

\begin{table*}
    \centering
    \caption{Reconstructions of blurry Gaussian-noisy images. (Stats over 256 samples.)}
    \begin{tabular}{ccccccc}
        $x$ & $\mbox{R}_\lambda(y^\delta)$ & $\mbox{R}_\lambda(G(y^\delta))$ & $\mbox{R}_\lambda(\mbox{BM3D}(y^\delta, \sigma))$ & $\mbox{R}_\lambda(\mbox{MEDIAN}(y^\delta))$ & $\mbox{R}_\lambda(G_{\text{supervised}}(y^\delta))$ & $\mbox{R}_\lambda(G_{\text{SURE}}(y^\delta, \sigma))$  \\
        \hline\\ [-2ex]

        \includegraphics[trim={9.53cm 4.06cm 59.46cm 3.4cm},clip, width=.12\textwidth]{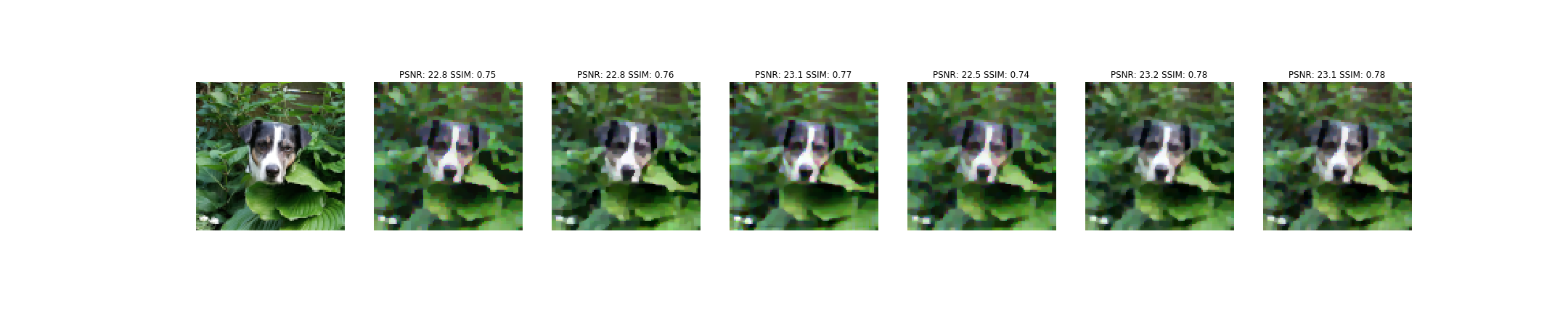} &
        \includegraphics[trim={18.17cm 4.06cm 50.81cm 3.4cm},clip, width=.12\textwidth]{figures/photos/3x3_Gaussian0.png} &
        \includegraphics[trim={26.82cm 4.06cm 42.17cm 3.4cm},clip, width=.12\textwidth]{figures/photos/3x3_Gaussian0.png} &
        \includegraphics[trim={35.46cm 4.06cm 33.52cm 3.4cm},clip, width=.12\textwidth]{figures/photos/3x3_Gaussian0.png} &
        \includegraphics[trim={44.11cm 4.06cm 24.88cm 3.4cm},clip, width=.12\textwidth]{figures/photos/3x3_Gaussian0.png} &
        \includegraphics[trim={52.75cm 4.06cm 16.23cm 3.4cm},clip, width=.12\textwidth]{figures/photos/3x3_Gaussian0.png} &
        \includegraphics[trim={61.4cm 4.06cm 7.59cm 3.4cm},clip, width=.12\textwidth]{figures/photos/3x3_Gaussian0.png} \\

        \includegraphics[trim={9.53cm 4.06cm 59.46cm 3.4cm},clip, width=.12\textwidth]{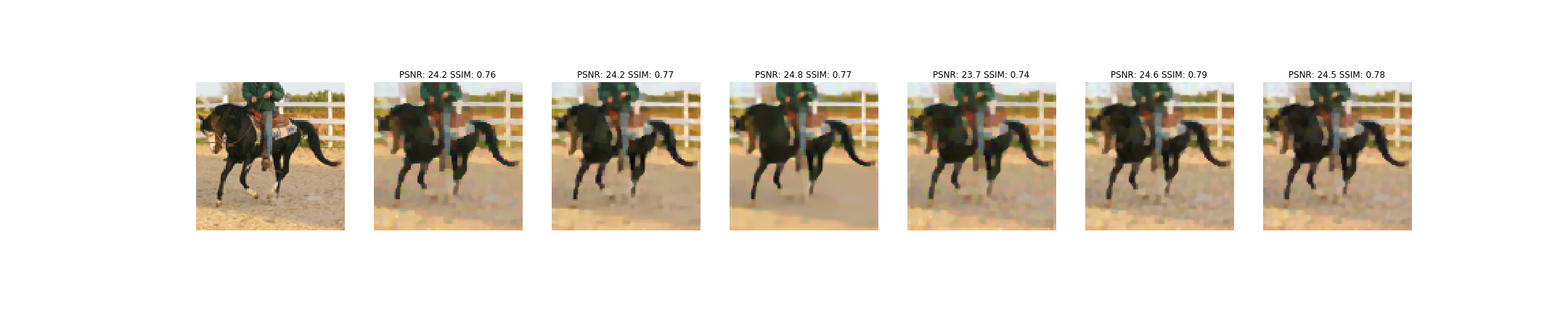} &
        \includegraphics[trim={18.17cm 4.06cm 50.81cm 3.4cm},clip, width=.12\textwidth]{figures/photos/3x3_Gaussian1.png} &
        \includegraphics[trim={26.82cm 4.06cm 42.17cm 3.4cm},clip, width=.12\textwidth]{figures/photos/3x3_Gaussian1.png} &
        \includegraphics[trim={35.46cm 4.06cm 33.52cm 3.4cm},clip, width=.12\textwidth]{figures/photos/3x3_Gaussian1.png} &
        \includegraphics[trim={44.11cm 4.06cm 24.88cm 3.4cm},clip, width=.12\textwidth]{figures/photos/3x3_Gaussian1.png} &
        \includegraphics[trim={52.75cm 4.06cm 16.23cm 3.4cm},clip, width=.12\textwidth]{figures/photos/3x3_Gaussian1.png} &
        \includegraphics[trim={61.4cm 4.06cm 7.59cm 3.4cm},clip, width=.12\textwidth]{figures/photos/3x3_Gaussian1.png} \\

        \includegraphics[trim={9.53cm 4.06cm 59.46cm 3.4cm},clip, width=.12\textwidth]{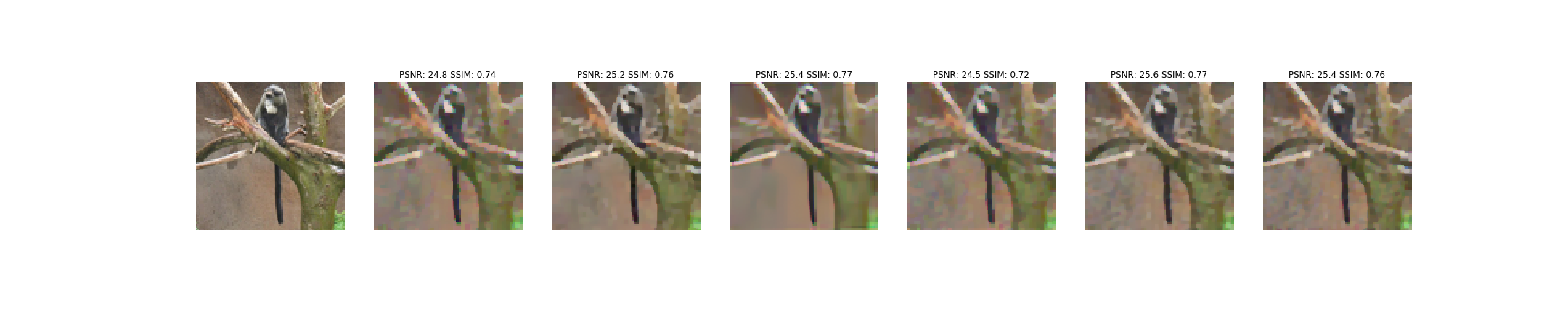} &
        \includegraphics[trim={18.17cm 4.06cm 50.81cm 3.4cm},clip, width=.12\textwidth]{figures/photos/3x3_Gaussian2.png} &
        \includegraphics[trim={26.82cm 4.06cm 42.17cm 3.4cm},clip, width=.12\textwidth]{figures/photos/3x3_Gaussian2.png} &
        \includegraphics[trim={35.46cm 4.06cm 33.52cm 3.4cm},clip, width=.12\textwidth]{figures/photos/3x3_Gaussian2.png} &
        \includegraphics[trim={44.11cm 4.06cm 24.88cm 3.4cm},clip, width=.12\textwidth]{figures/photos/3x3_Gaussian2.png} &
        \includegraphics[trim={52.75cm 4.06cm 16.23cm 3.4cm},clip, width=.12\textwidth]{figures/photos/3x3_Gaussian2.png} &
        \includegraphics[trim={61.4cm 4.06cm 7.59cm 3.4cm},clip, width=.12\textwidth]{figures/photos/3x3_Gaussian2.png} \\

        \hline
        \multirow{2}{*}{\textbf{PSNR stats}:}
        & \ \ mean:     $24.5$& \ \ mean:     $24.9$& \ \ mean:     $25.3$& \ \ mean:     $24.2$& \ \ mean:     $25.0$& \ \ mean:     $24.8$  \\
        & median:     $24.3$& median:     $24.9$& median:     $25.2$& median:     $24.0$& median:     $24.9$& median:     $24.7$  \\
         \hline
        \multirow{2}{*}{\textbf{SSIM stats}:}
        & \ \ mean:     $0.73$& \ \ mean:     $0.77$& \ \ mean:     $0.78$& \ \ mean:     $0.72$& \ \ mean:     $0.77$& \ \ mean:     $0.76$  \\
        & median:     $0.74$& median:     $0.78$& median:     $0.79$& median:     $0.73$& median:     $0.77$& median:     $0.77$  \\
    \end{tabular}
    \label{tab:results_3x3_Gaussian}
\end{table*}
\begin{table*}
    \centering
    \caption{Reconstructions of blurry Mixed-noisy images. (Stats over 256 samples.)}
    \begin{tabular}{ccccccc}
        $x$ & $\mbox{R}_\lambda(y^\delta)$ & $\mbox{R}_\lambda(G(y^\delta))$ & $\mbox{R}_\lambda(\mbox{BM3D}(y^\delta, \sigma))$ & $\mbox{R}_\lambda(\mbox{MEDIAN}(y^\delta))$ & $\mbox{R}_\lambda(G_{\text{supervised}}(y^\delta))$ & $\mbox{R}_\lambda(G_{\text{SURE}}(y^\delta, \sigma))$  \\
        \hline\\ [-2ex]

        \includegraphics[trim={9.53cm 4.06cm 59.46cm 3.4cm},clip, width=.12\textwidth]{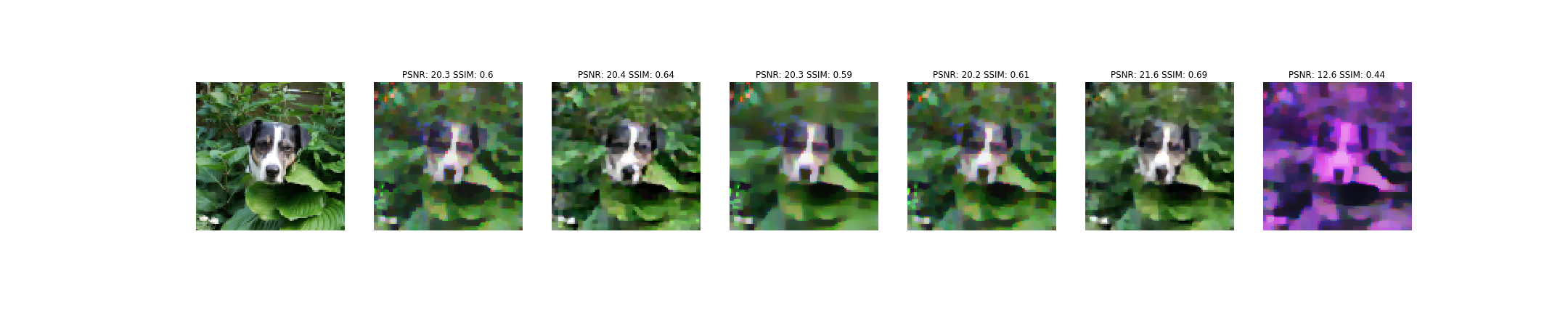} &
        \includegraphics[trim={18.17cm 4.06cm 50.81cm 3.4cm},clip, width=.12\textwidth]{figures/photos/3x3_Mixed0.png} &
        \includegraphics[trim={26.82cm 4.06cm 42.17cm 3.4cm},clip, width=.12\textwidth]{figures/photos/3x3_Mixed0.png} &
        \includegraphics[trim={35.46cm 4.06cm 33.52cm 3.4cm},clip, width=.12\textwidth]{figures/photos/3x3_Mixed0.png} &
        \includegraphics[trim={44.11cm 4.06cm 24.88cm 3.4cm},clip, width=.12\textwidth]{figures/photos/3x3_Mixed0.png} &
        \includegraphics[trim={52.75cm 4.06cm 16.23cm 3.4cm},clip, width=.12\textwidth]{figures/photos/3x3_Mixed0.png} &
        \includegraphics[trim={61.4cm 4.06cm 7.59cm 3.4cm},clip, width=.12\textwidth]{figures/photos/3x3_Mixed0.png} \\

        \includegraphics[trim={9.53cm 4.06cm 59.46cm 3.4cm},clip, width=.12\textwidth]{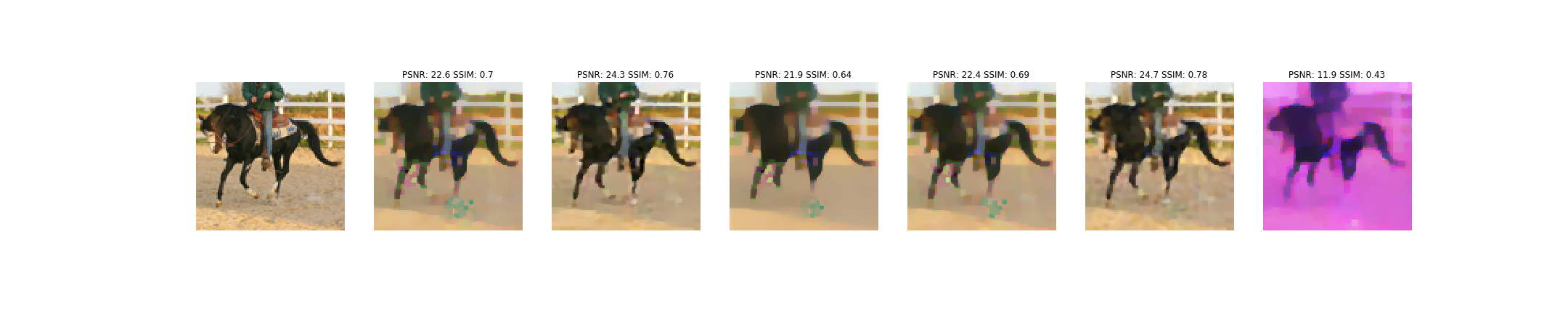} &
        \includegraphics[trim={18.17cm 4.06cm 50.81cm 3.4cm},clip, width=.12\textwidth]{figures/photos/3x3_Mixed1.png} &
        \includegraphics[trim={26.82cm 4.06cm 42.17cm 3.4cm},clip, width=.12\textwidth]{figures/photos/3x3_Mixed1.png} &
        \includegraphics[trim={35.46cm 4.06cm 33.52cm 3.4cm},clip, width=.12\textwidth]{figures/photos/3x3_Mixed1.png} &
        \includegraphics[trim={44.11cm 4.06cm 24.88cm 3.4cm},clip, width=.12\textwidth]{figures/photos/3x3_Mixed1.png} &
        \includegraphics[trim={52.75cm 4.06cm 16.23cm 3.4cm},clip, width=.12\textwidth]{figures/photos/3x3_Mixed1.png} &
        \includegraphics[trim={61.4cm 4.06cm 7.59cm 3.4cm},clip, width=.12\textwidth]{figures/photos/3x3_Mixed1.png} \\

        \includegraphics[trim={9.53cm 4.06cm 59.46cm 3.4cm},clip, width=.12\textwidth]{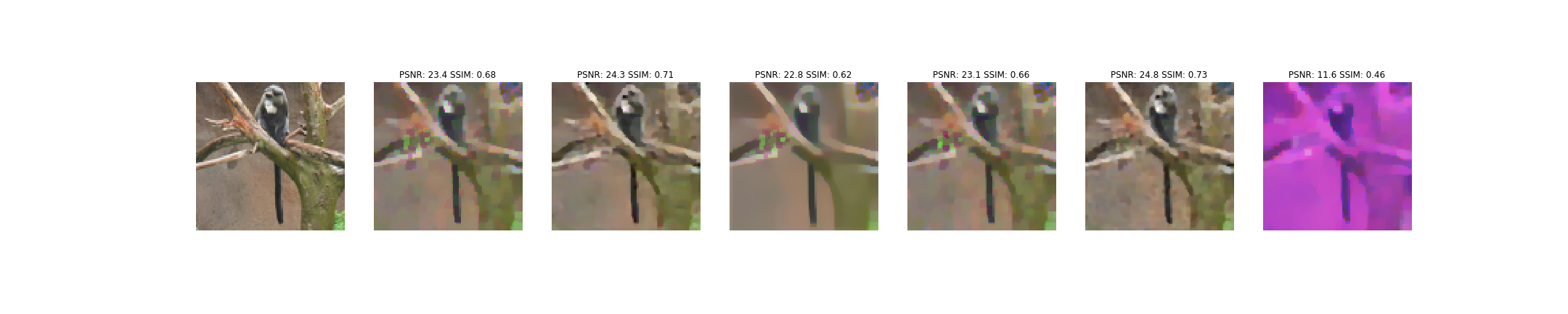} &
        \includegraphics[trim={18.17cm 4.06cm 50.81cm 3.4cm},clip, width=.12\textwidth]{figures/photos/3x3_Mixed2.png} &
        \includegraphics[trim={26.82cm 4.06cm 42.17cm 3.4cm},clip, width=.12\textwidth]{figures/photos/3x3_Mixed2.png} &
        \includegraphics[trim={35.46cm 4.06cm 33.52cm 3.4cm},clip, width=.12\textwidth]{figures/photos/3x3_Mixed2.png} &
        \includegraphics[trim={44.11cm 4.06cm 24.88cm 3.4cm},clip, width=.12\textwidth]{figures/photos/3x3_Mixed2.png} &
        \includegraphics[trim={52.75cm 4.06cm 16.23cm 3.4cm},clip, width=.12\textwidth]{figures/photos/3x3_Mixed2.png} &
        \includegraphics[trim={61.4cm 4.06cm 7.59cm 3.4cm},clip, width=.12\textwidth]{figures/photos/3x3_Mixed2.png} \\

        \hline
        \multirow{2}{*}{\textbf{PSNR stats}:}
        & \ \ mean:     $22.8$& \ \ mean:     $24.6$& \ \ mean:     $22.6$& \ \ mean:     $22.8$& \ \ mean:     $24.9$& \ \ mean:     $12.7$  \\
        & median:     $22.6$& median:     $24.3$& median:     $22.5$& median:     $22.6$& median:     $24.8$& median:     $12.7$  \\
         \hline
        \multirow{2}{*}{\textbf{SSIM stats}:}
        & \ \ mean:     $0.66$& \ \ mean:     $0.75$& \ \ mean:     $0.64$& \ \ mean:     $0.66$& \ \ mean:     $0.76$& \ \ mean:     $0.43$  \\
        & median:     $0.65$& median:     $0.77$& median:     $0.65$& median:     $0.66$& median:     $0.78$& median:     $0.43$  \\
    \end{tabular}
    \label{tab:results_3x3_Mixed}
\end{table*}
\begin{table*}
    \centering
    \caption{Reconstructions of blurry locally-noisy images. (Stats over 256 samples.)}
    \begin{tabular}{ccccccc}
        $x$ & $\mbox{R}_\lambda(y^\delta)$ & $\mbox{R}_\lambda(G(y^\delta))$ & $\mbox{R}_\lambda(\mbox{BM3D}(y^\delta, \sigma))$ & $\mbox{R}_\lambda(\mbox{MEDIAN}(y^\delta))$ & $\mbox{R}_\lambda(G_{\text{supervised}}(y^\delta))$ & $\mbox{R}_\lambda(G_{\text{SURE}}(y^\delta, \sigma))$  \\
        \hline\\ [-2ex]

        \includegraphics[trim={9.53cm 4.06cm 59.46cm 3.4cm},clip, width=.12\textwidth]{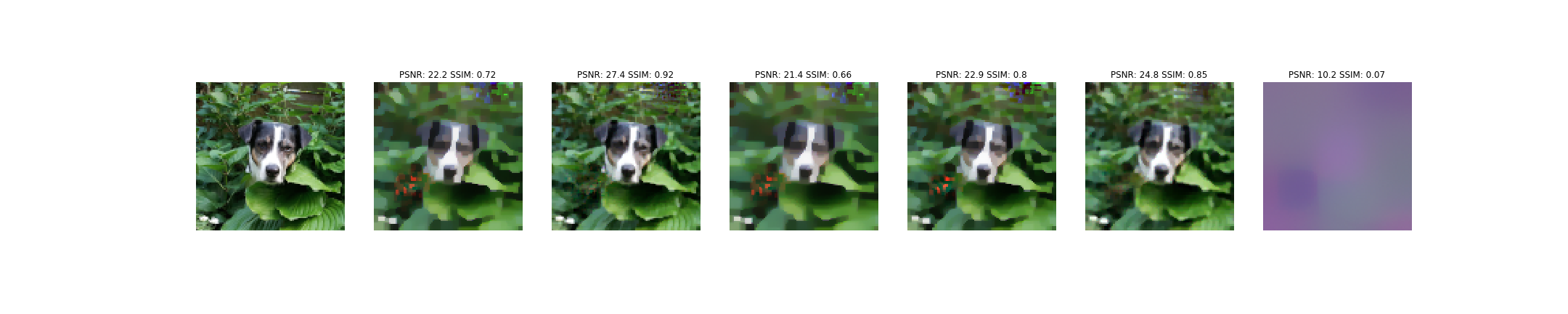} &
        \includegraphics[trim={18.17cm 4.06cm 50.81cm 3.4cm},clip, width=.12\textwidth]{figures/photos/3x3_Blob0.png} &
        \includegraphics[trim={26.82cm 4.06cm 42.17cm 3.4cm},clip, width=.12\textwidth]{figures/photos/3x3_Blob0.png} &
        \includegraphics[trim={35.46cm 4.06cm 33.52cm 3.4cm},clip, width=.12\textwidth]{figures/photos/3x3_Blob0.png} &
        \includegraphics[trim={44.11cm 4.06cm 24.88cm 3.4cm},clip, width=.12\textwidth]{figures/photos/3x3_Blob0.png} &
        \includegraphics[trim={52.75cm 4.06cm 16.23cm 3.4cm},clip, width=.12\textwidth]{figures/photos/3x3_Blob0.png} &
        \includegraphics[trim={61.4cm 4.06cm 7.59cm 3.4cm},clip, width=.12\textwidth]{figures/photos/3x3_Blob0.png} \\

        \includegraphics[trim={9.53cm 4.06cm 59.46cm 3.4cm},clip, width=.12\textwidth]{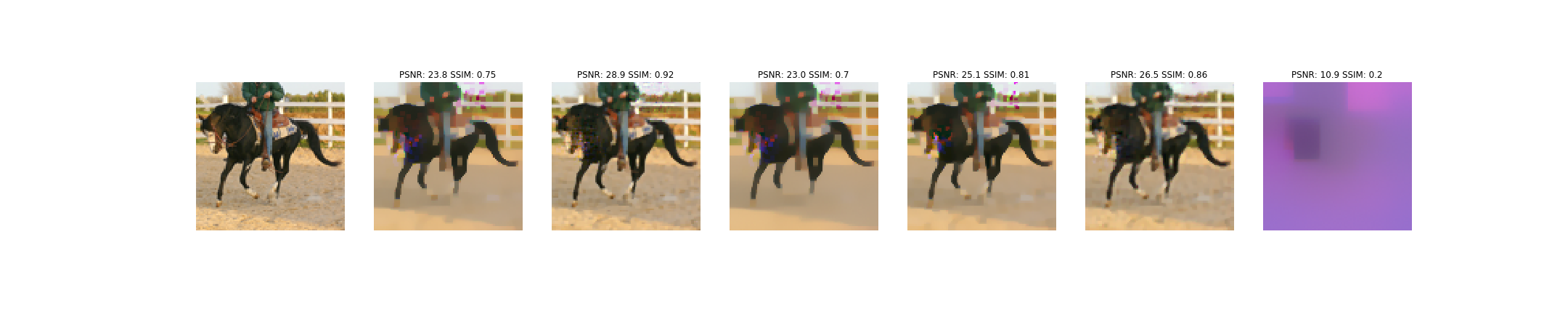} &
        \includegraphics[trim={18.17cm 4.06cm 50.81cm 3.4cm},clip, width=.12\textwidth]{figures/photos/3x3_Blob1.png} &
        \includegraphics[trim={26.82cm 4.06cm 42.17cm 3.4cm},clip, width=.12\textwidth]{figures/photos/3x3_Blob1.png} &
        \includegraphics[trim={35.46cm 4.06cm 33.52cm 3.4cm},clip, width=.12\textwidth]{figures/photos/3x3_Blob1.png} &
        \includegraphics[trim={44.11cm 4.06cm 24.88cm 3.4cm},clip, width=.12\textwidth]{figures/photos/3x3_Blob1.png} &
        \includegraphics[trim={52.75cm 4.06cm 16.23cm 3.4cm},clip, width=.12\textwidth]{figures/photos/3x3_Blob1.png} &
        \includegraphics[trim={61.4cm 4.06cm 7.59cm 3.4cm},clip, width=.12\textwidth]{figures/photos/3x3_Blob1.png} \\

        \includegraphics[trim={9.53cm 4.06cm 59.46cm 3.4cm},clip, width=.12\textwidth]{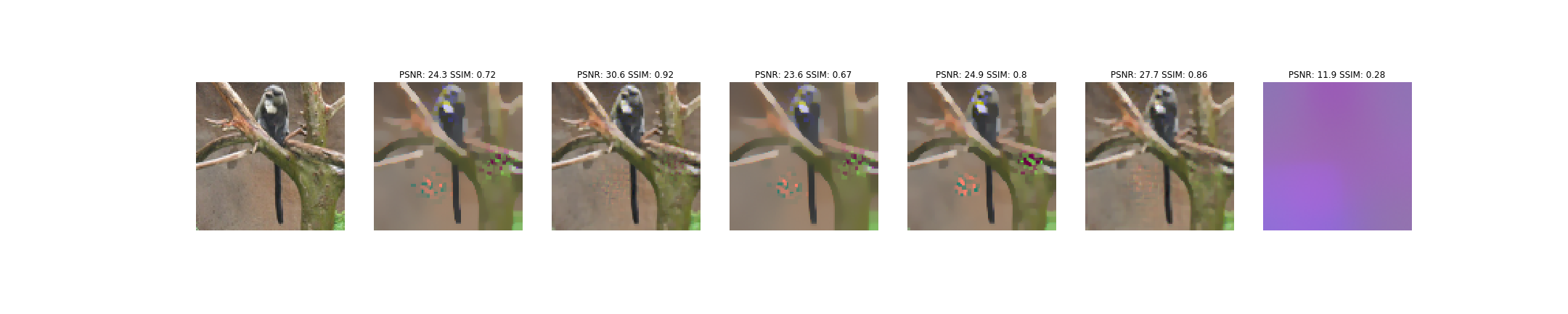} &
        \includegraphics[trim={18.17cm 4.06cm 50.81cm 3.4cm},clip, width=.12\textwidth]{figures/photos/3x3_Blob2.png} &
        \includegraphics[trim={26.82cm 4.06cm 42.17cm 3.4cm},clip, width=.12\textwidth]{figures/photos/3x3_Blob2.png} &
        \includegraphics[trim={35.46cm 4.06cm 33.52cm 3.4cm},clip, width=.12\textwidth]{figures/photos/3x3_Blob2.png} &
        \includegraphics[trim={44.11cm 4.06cm 24.88cm 3.4cm},clip, width=.12\textwidth]{figures/photos/3x3_Blob2.png} &
        \includegraphics[trim={52.75cm 4.06cm 16.23cm 3.4cm},clip, width=.12\textwidth]{figures/photos/3x3_Blob2.png} &
        \includegraphics[trim={61.4cm 4.06cm 7.59cm 3.4cm},clip, width=.12\textwidth]{figures/photos/3x3_Blob2.png} \\

        \hline
        \multirow{2}{*}{\textbf{PSNR stats}:}
        & \ \ mean:     $23.9$& \ \ mean:     $29.9$& \ \ mean:     $23.3$& \ \ mean:     $24.6$& \ \ mean:     $27.0$& \ \ mean:     $10.6$  \\
        & median:     $23.9$& median:     $30.0$& median:     $23.1$& median:     $24.7$& median:     $27.0$& median:     $10.7$  \\
         \hline
        \multirow{2}{*}{\textbf{SSIM stats}:}
        & \ \ mean:     $0.74$& \ \ mean:     $0.93$& \ \ mean:     $0.69$& \ \ mean:     $0.81$& \ \ mean:     $0.86$& \ \ mean:     $0.23$  \\
        & median:     $0.76$& median:     $0.93$& median:     $0.71$& median:     $0.82$& median:     $0.87$& median:     $0.2$  \\
    \end{tabular}
    \label{tab:results_3x3_localized_noise}
\end{table*}

\section{Generalization to multiplicative noise}
Before concluding in the next section, we would like to outline how our method can be applied to multiplicative noise and how a data set of noise could be accumulated.

We begin by discussing multiplicative noise. The whole method carries over mutatis mutandis to multiplicative noise, by making the observation that also for multiplicative noise the equation
\begin{equation}
    p_{y^\delta} = p_y * p_\eta
\end{equation}
holds, only from here on ``$*$'' denotes the convolution operator with respect to the multiplicative group -- whereas usually, and as used above, it refers to convolution with regard to the additive group. With this new way of reading ``$*$'', our observation 1, i.e., Equation~\eqref{eq:observation1}, carries over without any modification.
Our observation 2, i.e., Equation~\eqref{eq:observation2}, requires simple rewriting, reading 
\begin{equation}
\left(\id / G^*\right)_\#p_{y^\delta} = p_\eta,
\end{equation}
where the division is read to be point-wise. Note that in this case one has to take some care of the output of the generator to handle possible divisions by zero.

For collecting a noise data set in the multiplicative case, one could either utilize some known $y$ (or in the inverse problem case some known $x$), e.g., a 3D printed phantom, see~\cite{graeser2019, sedlacik2016magnetic}. Alternatively, one could make some assumptions on the noise, like zero-centeredness -- this assumption could also be used in the additive case if $y$ is not the result of a linear measurement.

We think our method of denoising could even be useful in some cases where the noise is known (or well approximated) in closed form, since -- unlike most other denoising methods and fidelity terms -- our approach incorporates not only information about $p_\eta$, but also about $p_y$ (via knowledge about $p_{y\delta}$).

\section{Conclusion}
This paper introduces a novel way of training a denoiser without any kind of ground truth data by making use of optimal transport, more specifically by using a modified Wasserstein GAN setting.
The method not only deals with non-Gaussian noise but even noise with spacial dependencies. It does so by not only incorporating information about the noise but also about the data distribution itself by using noisy data points of it. We also show that it rivals unsupervised state-of-the-art denoisers on Gaussian noise and outperforms them for other kinds of noise. Further, we show how the approach enables one to use deep learning methods to solve inverse problems without using any kind of ground truth data. We think the method has many use cases ranging from denoising of measurements and signals in general to more specific problems, like denoising seismic, audio data, or sinograms.

\FloatBarrier
\appendices
\section{}
\FloatBarrier

\begin{figure}[hbt!]
    \centering
    \includegraphics[trim={3cm 6cm 3cm 5cm},clip,width=.45\textwidth]{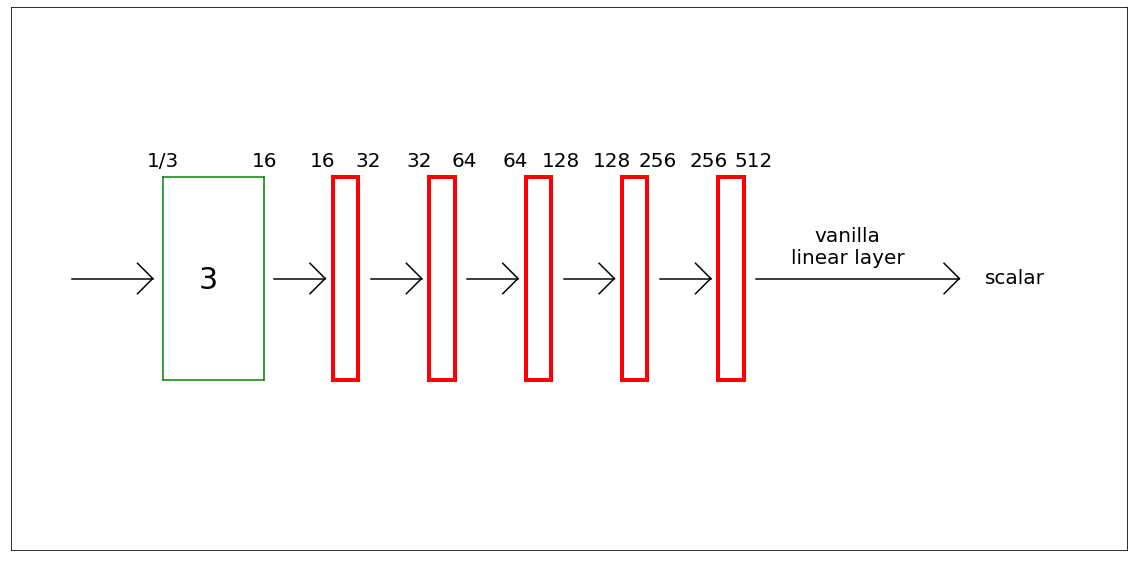}
    \caption{Critic architecture with a final linear dense layer. The green block is a linear convolutional layer, the red ones convolutional ResBlocks. The numbers above the blocks denote the number in-/output channels.}
    \label{fig:critic_architecture}
\end{figure}

\begin{figure}
    \centering
    \includegraphics[trim={2cm 4.5cm 2cm 4cm},clip,width=.47\textwidth]{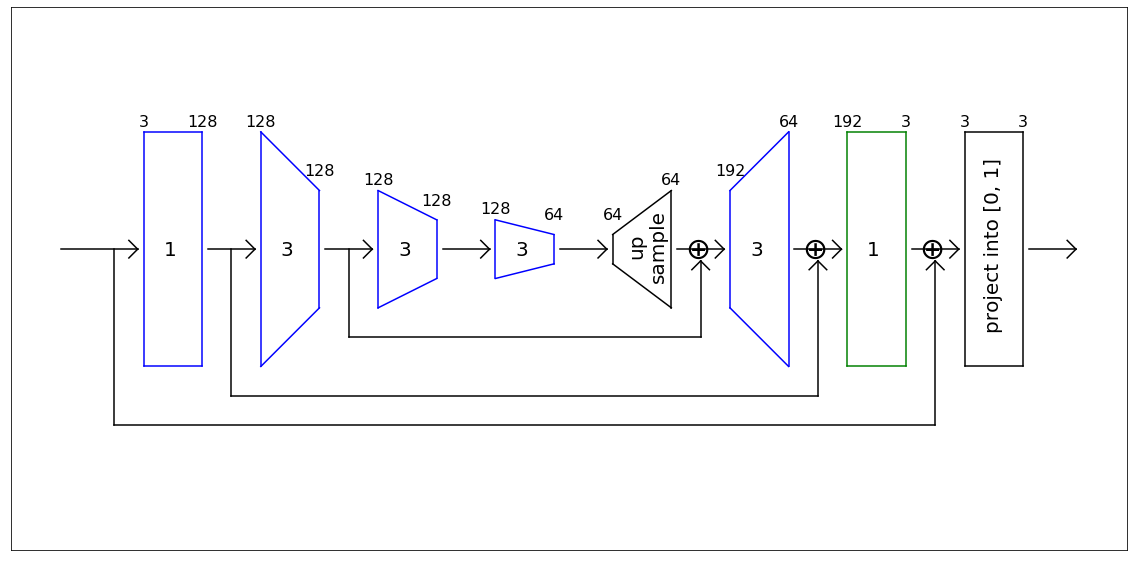}
    \caption{Image generator architecture. The color coding is analogous to Figures~\ref{fig:critic_architecture} and \ref{fig:1dim_generator_architecture}. I.e., blue denotes a convolution-layer-normalization-$\relu$ layer and green a linear convolution layer. The numbers within a block denotes the kernel size.}
    \label{fig:image_generator_architecture}
\end{figure}

\begin{figure}
    \centering
    \includegraphics[trim={2cm 2.5cm 2cm 1cm},clip,width=.47\textwidth]{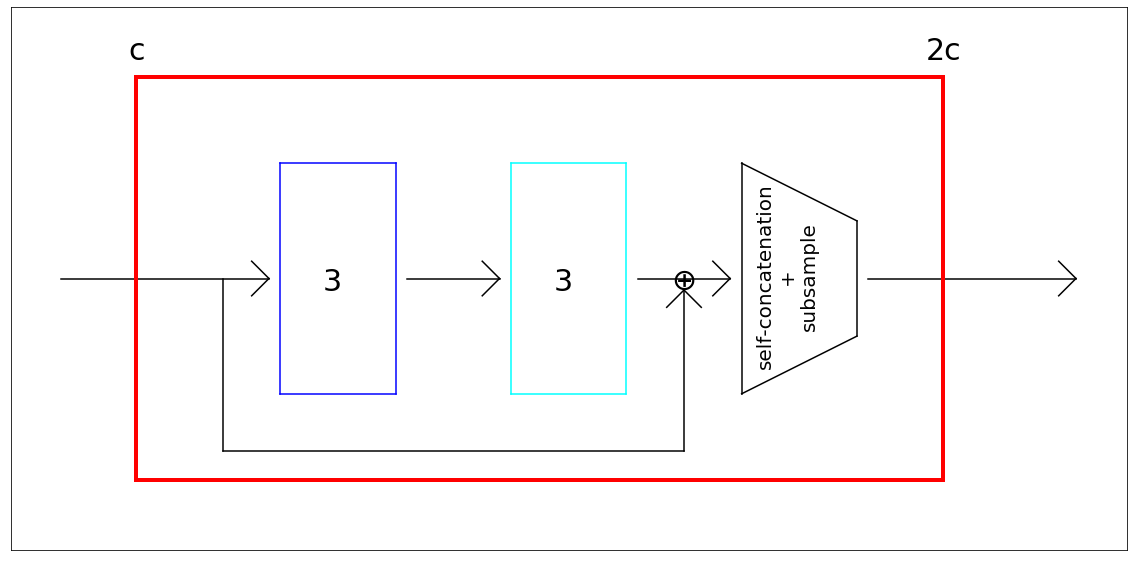}
    \caption{ResBlock interior. The blue block is a convolutional layer followed by a layer-normalization and a $\leakyrelu$ activation. The cyan block is convolutional layer followed by a layer-normalization. The trapezoid is a spacial-subsampling by the factor $2$ and a doubling of channels via self-concatenation of the tensor.}
    \label{fig:resBlock}
\end{figure}

\begin{figure}
    \centering
    \includegraphics[trim={2cm 6cm 2cm 5cm},clip,width=.47\textwidth]{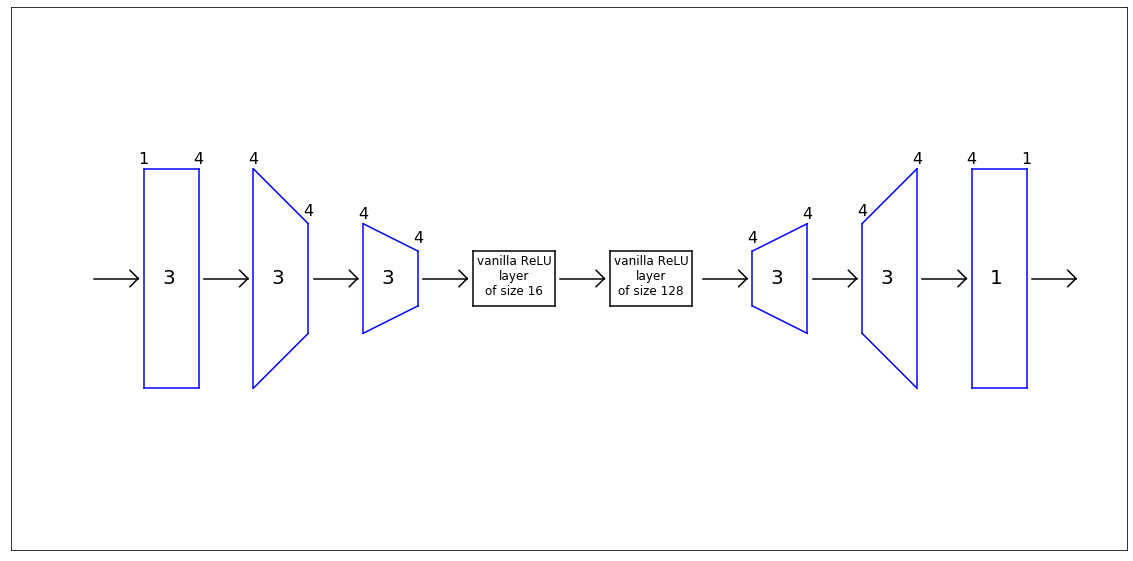}
    \caption{One-dimensional generator architecture.}
    \label{fig:1dim_generator_architecture}
\end{figure}
\newpage

\begin{table}
\tabcolsep=0.11cm
\centering
\caption{List of training parameters and details.}
\begin{tabular}{ccccc}
\hline
 Net/Task & Adam parameters & \rule{0pt}{1pt}\pbox[c]{20cm}{batch\\size} & \#batches & \rule{0pt}{1pt}\pbox[c]{5cm}{ training\\time\\in hours} \\
 \\ [-2ex]
 \hline
 \hline
 \rule{0pt}{15pt}\pbox[c]{20cm}{$1$-dim.\\ denoising} & (2e-4, 0.5, 0.9) & 8 & 6,750,000 & 125 \\ \\ [-2ex]
 \hline
 \rule{0pt}{15pt}\pbox[c]{20cm}{$1$-dim.\\SURE\\denoising} & (2e-4, 0.5, 0.9) & 8 & 30,330,000 & 125 \\ \\ [-2ex]
 \hline
 \rule{0pt}{15pt}\pbox[c]{20cm}{$1$-dim.\\supervised\\denoising} & (1e-4, 0.9, 0.999) & 8 & 43,850,000 & 125 \\ \\ [-2ex]
 \hline
 \rule{0pt}{15pt}\pbox[c]{20cm}{Image denoising\\Gaussian noise}  & (2e-4, 0.5, 0.9) & 152 & 1,550,000 & 750 \\ \\ [-2ex]
 \hline
 \rule{0pt}{15pt}\pbox[c]{20cm}{Image denoising\\Mixed noise}  & (2e-4, 0.5, 0.9) & 152 & 841,000 & 750 \\ \\ [-2ex]
 \hline
 \rule{0pt}{15pt}\pbox[c]{20cm}{Image Deblurring\\Gaussian noise}  & (2e-4, 0.5, 0.9) & 152 & 1,548,500 & 750 \\ \\ [-2ex]
 \hline
 \rule{0pt}{15pt}\pbox[c]{20cm}{Image Deblurring\\Mixed noise}  & (2e-4, 0.5, 0.9) & 152 & 863,000 & 750 \\ \\ [-2ex]
 \hline
 \rule{0pt}{15pt}\pbox[c]{20cm}{Supervised Image\\denoising\\Gaussian noise}  &  (1e-4, 0.9, 0.999) & 152 & 250,000 & 22 \\ \\ [-2ex]
 \hline
 \rule{0pt}{15pt}\pbox[c]{20cm}{Supervised Image\\denoising\\Mixed noise}  & (1e-4, 0.9, 0.999) & 152 & 200,000 & 28 \\ \\ [-2ex]
 \hline
 \rule{0pt}{15pt}\pbox[c]{20cm}{Supervised Image\\Deblurring\\Gaussian noise}  & (1e-4, 0.9, 0.999) & 152 & 250,000 & 22 \\ \\ [-2ex]
 \hline
 \rule{0pt}{15pt}\pbox[c]{20cm}{Supervised Image\\Deblurring\\Mixed noise}  & (1e-4, 0.9, 0.999) & 152 & 200,000 & 28 \\ \\ [-2ex]
 \hline
 \rule{0pt}{15pt}\pbox[c]{20cm}{SURE Image\\denoising\\Gaussian noise}  &  (2e-4, 0.9, 0.999) & 96 & \rule{0pt}{15pt}\pbox[c]{20cm}{\ \ \ \ \ 538,500\\(early stopping)} & 75 \\ \\ [-2ex]
 \hline
 \rule{0pt}{15pt}\pbox[c]{20cm}{SURE Image\\denoising\\Mixed noise}  & (2e-4, 0.9, 0.999) & 96 & 500,000 & 100 \\ \\ [-2ex]
 \hline
 \rule{0pt}{15pt}\pbox[c]{20cm}{SURE Image\\Deblurring\\Gaussian noise}  & (2e-4, 0.9, 0.999) & 96 & \rule{0pt}{15pt}\pbox[c]{20cm}{\ \ \ \ \ 193,000\\(early stopping)} & 27 \\ \\ [-2ex]
 \hline
 \rule{0pt}{15pt}\pbox[c]{20cm}{SURE Image\\Deblurring\\Mixed noise}  & (2e-4, 0.9, 0.999) & 96 & 500,000 & 100 \\ \\ [-2ex]
  \hline
\end{tabular}
\label{tab:parameters}
\end{table}

\FloatBarrier
%\subfile{toy_example_figure_appendix}
%\subfile{I_Gaussian_figure_appendix}
%\subfile{I_Mixed_figure_appendix}
%\subfile{I_Mixed_blobs_only_figure_appendix}
%\subfile{3x3_Gaussian_figure_appendix}
%\subfile{3x3_Mixed_figure_appendix}
%\subfile{3x3_Mixed_blobs_only_figure_appendix}

%\FloatBarrier
\section*{Acknowledgment}
S. Dittmer acknowledges funding by the Deutsche Forschungsgemeinschaft (DFG, German Research Foundation) - project number 281474342/GRK2224/1 ``Pi$^3$ : Parameter Identification - Analysis, Algorithms, Applications''.

C.-B. Sch\"onlieb acknowledges support from the Leverhulme Trust project on ‘Breaking the non-convexity barrier’, the Philip Leverhulme Prize, the EPSRC grants EP/S026045/1 and EP/T003553/1, the EPSRC Centre Nr. EP/N014588/1, the Wellcome Innovator Award RG98755, European Union Horizon 2020 research and innovation programmes under the Marie Skodowska-Curie grant agreement No. 777826 NoMADS and No. 691070 CHiPS, the Cantab Capital Institute for the Mathematics of Information and the Alan Turing Institute.

\bibliographystyle{IEEEtran}
\bibliography{IEEEabrv,references}

\end{document}